\documentclass{article}

    \usepackage[preprint]{neurips_2025}

\usepackage[utf8]{inputenc} 
\usepackage[T1]{fontenc}    
\usepackage{hyperref}       
\usepackage{url}            
\usepackage{booktabs}       
\usepackage{amsfonts}       
\usepackage{nicefrac}       
\usepackage{microtype}      
\usepackage{xcolor}         

\usepackage{amsmath}
\usepackage{amssymb}
\usepackage{mathtools}
\usepackage{amsthm}

\usepackage[capitalize,noabbrev]{cleveref}

\theoremstyle{plain}

\theoremstyle{definition}

\theoremstyle{remark}

\usepackage[textsize=tiny]{todonotes}

\usepackage{subfigure}
\usepackage{wrapfig}
\usepackage{graphicx}
\usepackage{subfigure}
\usepackage{multirow}
\usepackage{algorithm}
\usepackage{algorithmic}
\usepackage{pifont}

\title{Addressing the Collaboration Dilemma in Low-Data Federated Learning via Transient Sparsity}

\author{ Qiao Xiao$^{1}$, Boqian Wu$^{2, 3}$, Andrey Poddubnyy$^{1}$, Elena Mocanu$^{3}$, Phuong H. Nguyen$^{1}$, \and \textbf{Mykola Pechenizkiy$^{1}$, Decebal Constantin Mocanu$^{2,1}$}  \\
$^1$ Eindhoven University of Technology, $^2$ University of Luxembourg, $^3$ University of Twente \\
\texttt{\{q.xiao, a.poddubnyy, p.nguyen.hong, m.pechenizkiy\}@tue.nl}, \\
\texttt{\{b.wu, e.mocanu\}@utwente.nl}, \texttt{decebal.mocanu@uni.lu} \\
}

\begin{document}

\maketitle

\begin{abstract}
Federated learning (FL) enables collaborative model training across decentralized clients while preserving data privacy, leveraging aggregated updates to build robust global models. However, this training paradigm faces significant challenges due to data heterogeneity and limited local datasets, which often impede effective collaboration.
In such scenarios, we identify the Layer-wise Inertia Phenomenon in FL, wherein the middle layers of global model undergo minimal updates after early communication rounds, ultimately limiting the effectiveness of global aggregation. We demonstrate the presence of this phenomenon across a wide range of federated settings, spanning diverse datasets and architectures. To address this issue, we propose LIPS (Layer-wise Inertia Phenomenon with Sparsity), a simple yet effective method that periodically introduces \textit{transient sparsity} to stimulate meaningful updates and empower global aggregation.
Experiments demonstrate that LIPS effectively mitigates layer-wise inertia, enhances aggregation effectiveness, and improves overall performance in various FL scenarios. This work not only deepens the understanding of layer-wise learning dynamics in FL but also paves the way for more effective collaboration strategies in resource-constrained environments. Our code is publicly available at: \url{https://github.com/QiaoXiao7282/LIPS}.


\end{abstract}

\section{Introduction}
\label{introduction}

Federated learning (FL) \citep{mcmahan2016federated, yang2019federated} enables collaborative training of machine learning models across decentralized clients while keeping the raw data local, making it a widely used solution for addressing privacy concerns and data access limitations in domains such as healthcare \citep{rieke2020future, sadilek2021privacy}, finance \citep{long2020federated}, and personalized services \citep{long2020federated, wen2023survey}. For example, in medical applications, where patient data is both scarce and highly sensitive, FL enables the collaborative training of robust models without compromising privacy \citep{nguyen2022federated, rieke2020future}.


However, a significant challenge in FL lies in the non-independent and identically distributed (non-IID) nature of data across clients \citep{zhao2018federated, li2022federated}. Such data heterogeneity often leads to divergent local updates, making it difficult to train a global model that generalizes well across all clients. To address this issue, various methods have been proposed, including personalizing models to better adapt to local data distributions and designing advanced aggregation mechanisms to alleviate the challenges posed by non-IID scenarios \citep{li2020federated, tan2022towards, wu2023bold, tamirisa2024fedselect}.

Among these approaches, layer-wise aggregation methods have gained increasing attention by selectively aggregating specific layers while leaving others client-specific \citep{collins2021exploiting, ma2022layer, zhang2023fedala}. 
For instance, FedBN \citep{lifedbn} performs global aggregation except batch normalization layers, keeping them client-specific to mitigate the impact of non-IID data, while FedRep \citep{collins2021exploiting} similarly excludes the classifier to better align with local tasks.
These methods inherently assume that the remaining aggregated layers are capable of supporting meaningful collaboration across clients. However, this assumption may not hold—particularly in low-data regimes, where each client has access to only a limited amount of local training data, exacerbating statistical heterogeneity and increasing the risk of overfitting.
The efficacy of aggregation in FL, particularly at the layer level under low-data regimes, has not been thoroughly investigated, leaving a significant gap in understanding how layers behave under such constrained scenarios.  

Having said that, the collaboration through aggregation in FL \citep{mcmahan2016federated, yang2019federated}, which pools insights from diverse clients with access to broader data should, in theory, address the overfitting risks posed by limited local data. However, our investigation reveals a startling paradox: far from resolving overfitting, the collaborative process in FL perpetuates it, manifesting as the \textit{Layer-wise Inertia Phenomenon}, where intermediate layers of the global model exhibit little to no meaningful updates after the early communication rounds. This stagnation limits the effectiveness of aggregation, indicating that not all aggregated layers contribute meaningfully to the global model after aggregation. 
What's worse, the layer-wise inertia phenomenon intensifies in deeper models and persists across diverse data distributions and client numbers, undermining FL’s core promise of effective collaboration. We refer to this limitation as the \textit{collaboration dilemma} in FL. 



To address this issue, we propose Layer-wise Inertia Phenomenon with Sparsity (LIPS), a novel method designed to enhance layer-wise aggregation in FL. By periodically introducing transient sparsity, which temporarily deactivating parameters with low sensitivity after each communication round, LIPS redistributes the model's learning capacity, fostering more effective updates and enabling the global model to perform better across all clients. 
We evaluate LIPS across diverse architectures and datasets, including CIFAR-10 \citep{krizhevsky2010cifar}, CIFAR-100 \citep{krizhevsky2009learning}, and TinyImageNet \citep{le2015tiny}, using VGG \citep{simonyan2014very} and ResNet-based \citep{he2016deep} models. 
Our contributions in this work can be summarized as follows: 

\begin{itemize}
    \item We identify and demonstrate the existence of the layer-wise inertia phenomenon in low-data federated learning, where certain layers exhibit early stagnation.
\end{itemize}

\begin{itemize}
    \item We analyze the underlying causes of this phenomenon and highlight its negative impact on learning dynamics and its constraint on the overall performance of federated learning.
\end{itemize}

\begin{itemize}
    \item We propose LIPS, a simple yet effective method that introduces transient sparsity to mitigate the inertia phenomenon and enhance the aggregation during federated training.
\end{itemize}

\begin{itemize}
    \item We validate the effectiveness of LIPS in enhancing model performance and promoting more effective collaboration across a variety of federated learning scenarios.
\end{itemize}


\section{Preliminary}

\textbf{Federated learning.} Federated learning (FL) strives to enable collaborative training of deep models across decentralized clients while preserving data privacy by keeping local data $\mathcal{D}_i$ on devices. A global model is iteratively trained using local updates from multiple clients. One of the commonly used strategies in FL, FedAvg \citep{mcmahan2017communication}, aims to train a shared model by allowing each client to update a local model using its own data. Periodically, clients send their updated parameters to a central server, which aggregates these updates (typically via averaging) and redistributes the aggregated model back to the clients for further refinement. This approach enhances privacy and scalability compared to traditional distributed learning methods.
The global objective is typically formulated as:

\begin{equation}
\min _{\mathbf{w}} F(\mathbf{w})=\sum_{i=1}^n \frac{\left|\mathcal{D}_i\right|}{|\mathcal{D}|} F_i(\mathbf{w})
\label{fedavg}
\end{equation}

where $F_i(\mathbf{w})=\frac{1}{\left|\mathcal{D}_i\right|} \sum_{(x, y) \in \mathcal{D}_i} \mathcal{L}(f(\mathbf{w} ; x), y)$ represents the local loss for client $i$, and $F(\mathbf{w})$ is the global objective. Here, \(\mathcal{D}_i\) denotes the local dataset for client \(i\)  and \(|\mathcal{D}_i|\) is the number of samples in the dataset of client \(i\). The global dataset \(\mathcal{D}\) is the union of all clients’ datasets, i.e., \(\mathcal{D} = \bigcup_{i=1}^n \mathcal{D}_i\), and \(|\mathcal{D}|\) is the total number of samples across all clients. The term $\mathcal{L}$ is a loss function (e.g., cross-entropy), and $f(\mathbf{w} ; x)$ is the model's prediction.

Despite its success in classical FL tasks, FedAvg often struggles in non-IID settings, where heterogeneous data across clients causes slow convergence and reduced accuracy. Recent approaches address this by localizing parameters sensitive to non-IID data \citep{collins2021exploiting, lifedbn, zhang2023fedala}. By selectively sharing and localizing layers, these methods effectively mitigate the impact of data heterogeneity, improving FL performance.

These observations highlight that different layers contribute unevenly to global aggregation, particularly under non-IID conditions. This motivates a deeper exploration of layer-wise learning dynamics in FL, which remain largely underexplored, especially in low-data regimes where limited local examples further exacerbate the challenge.


\textbf{Layer-wise cosine similarity (global model).} 
To investigate how individual layer of the global model evolve over communication rounds in FL, we adopt layer-wise cosine similarity as a measurement metric—commonly used in prior work to assess the stability or drift of neural network weights or outputs over time (e.g., \citep{chen2025streamlining, gromov2025the, min2025docs, jiang2025tracing}). This metric enables us to quantify the extent of change in each layer's parameters, providing insight into learning dynamics across the network depth.

Specifically, for a given layer $l$ in the global model, we define the cosine similarity $C_l^t$ between the layer's weights at communication round $t$ and an earlier reference round $t_0$ as follows:
\begin{equation}
C_l^t=\frac{\mathbf{w}_l^t \cdot \mathbf{w}_l^{t_0}}{\left\|\mathbf{w}_l^t\right\|\left\|\mathbf{w}_l^{t_0}\right\|}
\end{equation}
where $\mathbf{w}_l^t$ and $\mathbf{w}_l^{t_0}$ are the weight vectors of $l$-th layer in the global model at round $t$ and ${t_0}$, respectively. A value of $C_l^t$ close to 1 indicates that the layer has undergone little change, while a smaller value reflects more substantial updates. 

\section{Layer-wise Learning Behavior in Low-Data FL}
\label{inertia_phenomenon}

To better understand how different layers of the global model behave during training in low-data federated learning (FL), we conduct experiments on the CIFAR-10, CIFAR-100, and TinyImageNet datasets using standard FedAvg algorithm, modified to keep batch normalization (BN) layers local to each client. This adjustment, inspired by \citep{lifedbn}, mitigates the negative effects of distribution shifts and improves the stability of the model in heterogeneous settings. Each client is assigned 100 training samples to simulate a low-data regime.
Further implementation details are provided in Section \ref{empirical_evaluations}. 

\begin{figure*}[!htb]
    \centering
    \includegraphics[width=\textwidth]{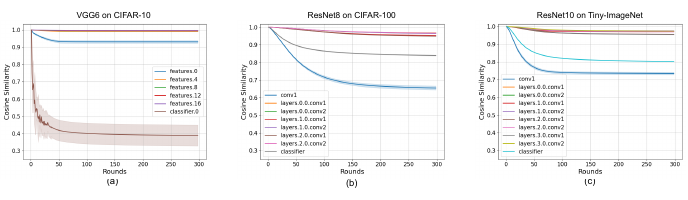}
    \vskip -0.15in
    \caption{Layer-wise cosine similarity of global model throughout the training process. We track the global model's layer-wise cosine similarity relative to the 2nd communication round states after aggregation, across the training process on CIFAR-10, CIFAR-100, and Tiny-ImageNet datasets using VGG6, ResNet-8, and ResNet-10 architectures, respectively, as shown in (a), (b), and (c). The number of clients is set to 100, with each client having 100 training samples for CIFAR-10 and CIFAR-100, and 300 training samples for Tiny-ImageNet, under a Dirichlet data distribution with Dir($\alpha$=0.1). The legend indicates the layer names for each architecture.}
    \vskip -0.1in
    \label{fig: fd_inertia}
\end{figure*}

To capture how individual layers evolve during training, we track the layer-wise cosine similarity of the global model after each aggregation step, computed relative to its state in an early communication round (e.g., $t_0 = 2$). This allows us to assess how each layer evolves over time. Our analysis reveals several key insights into layer-wise learning dynamic in low-data FL, discussed in detail below:

\textbf{{\large \ding{172}} In low-data FL, the global model exhibits the Layer-wise Inertia Phenomenon, wherein updates stagnate early and hinder further adaptation.}
As shown in Figure~\ref{fig: fd_inertia}, we uncover an intriguing phenomenon: most layers, particularly the middle layers, exhibit minimal changes during the training process, after the early phases of training. This is evidenced by the consistently high cosine similarity values (above 0.95) overall of the middle layers' weights when compared to their states in the early training phase. Notably, this behavior is consistently observed across different datasets and architectures. Additional results for other settings can be found in Appendix~\ref{app:cosine}.

We term this phenomenon \textbf{Layer-wise Inertia} in FL, describing the tendency of certain layers to experience stagnation, with minimal or slow updates during aggregation. This stagnation implies that the majority of layers contribute little to the overall model updates after the early training phases, leaving only a few layers (e.g., the first and last) as the primary drivers of collaboration. As a result, the aggregation becomes less effective. We believe that understanding and addressing this inertia has the potential to enhance learning dynamics and significantly improve the global performance of federated models.

\begin{figure}[!htb]
    \centering
    \includegraphics[width=\textwidth]{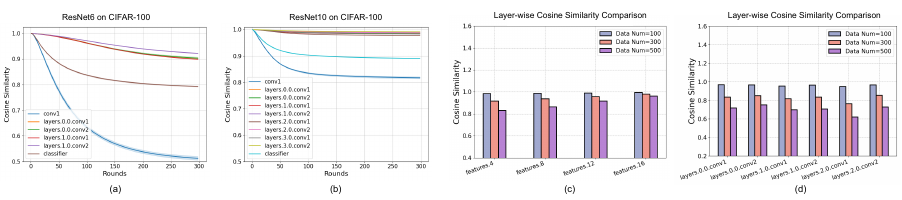}
    \vskip -0.1in
    \caption{Layer-wise cosine similarity of the global model after aggregation. (a) and (b): Results on CIFAR-100 with 100 clients using ResNet6 and ResNet10 under Dir($\alpha$=0.1). (c) and (d): Comparison under different data volumes per client (100, 300, 500 samples) on CIFAR-10 and CIFAR-100.}
    \label{fig:fd_datamodel}
    \vskip -0.2in
\end{figure}

We further observe that layer-wise inertia becomes more pronounced as model depth increases (Figure~\ref{fig:fd_datamodel} (a), (b)) or as the amount of training data per client decreases (Figure~\ref{fig:fd_datamodel} (c), (d)). This escalation suggests that the observed layer-wise stagnation is primarily attributed to overfitting, which diminishes the variability of local updates transmitted to the server, thereby reducing the diversity of information integrated during global aggregation and ultimately driving the stagnation of specific layers. We further compare this layer-wise behavior with centralized training under low-data regimes in Appendix~\ref{app:centralize}.
These insights underscore a critical limitation in FL: far from overcoming overfitting through collaborative aggregation, FL struggles to prevent layer stagnation for middle layers, posing significant challenges to scalability and generalization.

\textbf{{\large \ding{173}} Layer-wise Inertia limits the collaborative effectiveness of aggregation in the global model.}
We might ask: \textit{How does layer-wise inertia in the middle layers impact aggregation effectiveness?}
To investigate this, we conduct an experiment where the first and last layers of the model are aggregated throughout training, while the weights of middle layers remain fixed on each client after certain communication rounds. For simplicity, we set this point at round 50 for all datasets, based on observations in Figure \ref{fig: fd_inertia}, where most layers tend to exhibit minimal updates within clients beyond this point. We term this approach $\mathtt{w/.\hspace{0.2em}fix}$, while the standard approach, where the entire model is aggregated throughout training, is termed $\tt w/o.\hspace{0.2em}fix$.

\begin{wraptable}{r}{0.55\textwidth}
\vskip -0.2in
\begin{minipage}{0.55\textwidth}
    \centering
    \caption{Comparison of test accuracy on CIFAR-10, CIFAR-100, and TinyImageNet in federated learning, with and without fixing middle layer aggregation after the early stages of communication.}
    \label{tab:without_agg}
    \resizebox{\linewidth}{!}{
    \begin{tabular}{@{}lccc@{}}
        \toprule
        \textbf{Dataset}  & CIFAR-10 & CIFAR-100 & Tiny ImageNet \\ 
        \midrule
        $\tt w/o.\hspace{0.2em}fix$         & 85.83 $\pm$ 1.28 & 43.08 $\pm$ 0.45 &  36.83 $\pm$ 0.22  \\
        $\tt w/.\hspace{0.2em}fix$ &        85.15 $\pm$ 1.59  & 43.07  $\pm$ 0.53 & 36.86  $\pm$ 0.08    \\ 
        \bottomrule
    \end{tabular}
    }
\end{minipage}
\vskip -0.1in
\end{wraptable}

As shown in Table~\ref{tab:without_agg}, the performance under the $\tt w/.\hspace{0.2em}fix$ approach is surprisingly comparable to that achieved with $\tt w/o.\hspace{0.2em}fix$. This suggests that middle layer aggregation contributes minimally after the early stages of communication. This can be attributed to the minimal updates in the middle layers, which significantly restrict collaborative updates from clients and diminish the overall effectiveness of aggregation.
These findings suggest that overcoming layer-wise inertia, especially in the middle layers, may be key to unlocking more effective collaboration and improving model performance in low-data FL.

\begin{wrapfigure}{r}{0.5\textwidth}
\begin{minipage}{0.5\textwidth}
    \centering
    \vskip -0.1in
    \includegraphics[width=\textwidth]{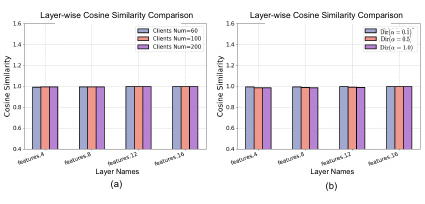}
    \vskip -0.1in
    \caption{Comparison of the global model's layer-wise cosine similarity after training on CIFAR-10. (a) Varying number of clients and (b) under different data distributions.}
    \vskip -0.1in
    \label{fig: fd_dist}
\end{minipage}
\end{wrapfigure}

\textbf{{\large \ding{174}} Client diversity across data distributions and client numbers struggles to solve the Layer-wise Inertia Phenomenon.}
To investigate whether standard configuration adjustments in FL can mitigate this stagnation, we examine the layer-wise cosine similarity $C_l^T$ under varying numbers of clients and different levels of data heterogeneity (via Dirichlet distributions), as shown in Figure~\ref{fig: fd_dist}. 
Across all settings, we consistently observe higher cosine similarity values, 
indicating that the final states of the middle layers in the global model 
remain closer to their early states, regardless of the data distribution or the number of clients. 
These results suggest that increasing client heterogeneity, either by altering the number of clients or the non-IID level, does not meaningfully mitigate the stagnation of specific layers. 
This highlights a fundamental limitation in low-data FL, where standard sources of variability are insufficient to promote more effective collaboration across all layers. This underscores the importance of explicitly addressing layer-wise inertia to revitalize stale layers and enhance the overall effectiveness of collaboration through aggregation in low-data FL.


\section{Solving Layer-wise Inertia Phenomenon with Sparsity}
\label{methods}

To address the underlying cause of the Layer-wise Inertia Phenomenon, we propose LIPS (Layer-wise Inertia Phenomenon with Sparsity), a simple yet effective approach that introduces transient sparsity, temporarily applied after aggregation to encourage more meaningful updates to the global model. The overall procedure is outlined in Algorithm~\ref{alg:LIPS}.

\begin{wrapfigure}{r}{0.56\textwidth}
\vskip -0.3in
\begin{minipage}{0.56\textwidth}
\begin{algorithm}[H]
	\caption{LIPS}
	\label{alg:LIPS}
	{\small
		\begin{algorithmic}
			\STATE {\bfseries Input:} Each client's initial model $\mathbf{w}_i^0$;
			Number of clients $n$;
			Total communication round $T$; Local epoch number $E$;
			Hyperparameters $\tau_0$, $k$.
			\STATE {\bfseries Output:} Model $\mathbf{w}_i^T$ for each client. \\
			\FOR {$t = 1$ to $T$} 
            
                \STATE \textcolor{blue}{\textbf{Client-side:}}
                 \IF {$t \bmod k = 0$}
                    \STATE Calculate sparsity based on $\tau \left(t ; \tau_0, T\right)$.
                    \STATE Evaluate the sensitivity of each parameter by Eq~. \eqref{sensity}.
                    \STATE Obtain the mask matrix $M_i^t$ and apply sparsification.
                \ENDIF
    		\FOR{$i=1$ to $N$ \textbf{in parallel}}
                    \STATE Update $\mathbf{w}_i^t$ for $E$ local epochs.
    		\ENDFOR
                \STATE Send $\mathbf{w}_i^t$ to the server.
            
                \STATE \textcolor{blue}{\textbf{Server-side:}}
                \STATE Compute the global model $\overline{\mathbf{w}}^t$ by average aggregation .
                \STATE Send $\overline{\mathbf{w}}^t$ to each client $i$.
                
                \STATE \textcolor{blue}{\textbf{Client-side:}}
                \FOR{$i=1$ to $N$ \textbf{in parallel}}
                    \STATE Initialize $\mathbf{w}_i^{t+1}$ with $\overline{\mathbf{w}}^t$, excluding BN layers.
                \ENDFOR
		\ENDFOR
	\end{algorithmic}}
\end{algorithm}
\end{minipage}
\vskip -0.2in
\end{wrapfigure}

\subsection{Why Use Sparsity}

Based on our observations in Section \ref{inertia_phenomenon}, the Layer-wise Inertia phenomenon on global model is primarily driven by overfitting, which leads to minimal updates during training. To address this, we propose leveraging sparsity as a means to actively reshape the learning dynamics. By removing a fraction of weights, the model is encouraged to redistribute its capacity and explore alternative optimization trajectories that lead to more meaningful and effective updates. This is motivated by principles from the sparse training literature \citep{evci2022gradient, chen2022sparsity}, where introducing sparsity has been shown to enhance generalization and improve optimization behavior. 

More specifically, this disruption by introducing sparsity breaks reliance on saturated weights, promotes greater variability in local updates, and revitalizes otherwise stagnant layers, thereby fostering a healthier learning dynamic in low-data FL. What’s more, the proposed approach is model-agnostic, easy to implement, and computationally lightweight, requiring no additional modules or retraining pipelines, making it a practical drop-in solution for real-world FL scenarios.

\subsection{Where to Sparsify: Sensitivity-Guided Parameter Selection}

A crucial component of LIPS is the selection of parameters to be zeroed out after model aggregation. To achieve this, LIPS employs a \textbf{sensitivity-based criterion} to identify parameters that are less critical to the model's performance at the local client level. Sensitivity, as discussed in prior research~\citep{lee2019snip, wu2023bold, nowak2024fantastic}, quantifies the importance of a parameter based on its impact on the model’s output or loss function when set to zero. Parameters with low sensitivity have limited influence on learning and are more likely to produce meaningless updates, especially under limited data where overfitting is prevalent. As such, they become ideal candidates for removing, encouraging more effective and meaningful updates in subsequent training rounds.

In LIPS, the sensitivity of the $j$-th parameter \( w_{i,j}^t \) in client \( i \) at time step \( t \) is calculated using the following metric:

\vskip -0.1in
\begin{equation}
s_{i,j}^t = \left| \Delta w_{i,j}^t \cdot w_{i,j}^{t} \right|,
\label{sensity}
\end{equation}
\vskip -0.1in

where \( \Delta w_{i,j}^t \) represents the update of the parameter \( w_{i,j} \) during training at round \( t \).

This metric captures the contribution of a parameter to the model’s updates, reflecting its importance in the learning process. Parameters with lower sensitivity \( s_{i,j}^t \) are prioritized for being zeroed out. A detailed derivation of this criterion is provided in Appendix~\ref{appendix:sensitivity}, with further discussion and comparison to other criteria presented in Section \ref{sec:ablation}.

\subsection{How to Sparsify: Periodic Transient Sparsity}

After evaluating parameter sensitivity using the metric in Eq.~\eqref{sensity}, LIPS temporarily zeros out \(\tau\) fraction of parameters with the lowest sensitivity in each layer for client \( i \). This process is implemented using a binary mask matrix \( M_i^t \), which determines whether a parameter is retained or set to zero for client $i$ at time step $t$. The elements of the mask are defined as follows:
\vskip -0.1in
\begin{equation}
m_{i,j}^t = 
\begin{cases} 
0, & \text{if } s_{i,j}^t \text{ is among the lowest} \tau \text{ fraction of in its layer}, \\
1, & \text{otherwise}.
\end{cases}
\end{equation}
\vskip -0.1in
The sparsity ratio \( \tau \) is decayed over time to allow more parameters to be retained in later rounds, promoting convergence stability. The decay function for $\tau$ is defined as: $\tau(t; \tau_0, T) = \tau_0 \left(1 - {t}/{T}\right)$, where $\tau_0$ is the initial sparsity ratio, $T$ is the total number of communication rounds.
Then the updated weight matrix for client \( i \) is obtained by: $\mathbf{w}_i^t = M_i^t \odot \mathbf{w}_i^t$, producing a sparsified model used as the starting point for local training.

The mask is applied periodically every $k$ communication rounds to avoid excessive disruption to model convergence. Notably, when sparsification is triggered, the mask is applied only once after global aggregation and is lifted before local training begins. As a result, local training still proceeds in a dense fashion. We refer to this mechanism as \textit{transient sparsity}. By periodically introducing transient sparsity, the model is encouraged to redistribute its learning capacity, stimulating underutilized parameters and mitigating stagnation in layers. 
This temporary application of sparsity allows the model to benefit from improved learning dynamics without permanently reducing its capacity, thereby preserving the full expressive power during local training.
To assess its effectiveness, we compare this design with maintaining sparsity during local training, as discussed in the Appendix~\ref{app:sparse_training}.

Based on our observations, the layer-wise inertia phenomenon is most pronounced in the middle layers of the model. Consequently, we apply sparsity only to the middle layers, excluding the first and last layers from sparsification. 


\section{Experiments}
\label{empirical_evaluations}

\textbf{Dataset and architectures.} We conduct experiments on three datasets, including CIFAR-10 \citep{krizhevsky2010cifar}, CIFAR-100 \citep{krizhevsky2009learning}, and TinyImageNet \citep{le2015tiny}. To evaluate the effectiveness of our method in different scenarios, we use the commonly adopted Dirichlet non-IID setting \citep{hsu2019measuring, lin2020ensemble, wu2023bold}, where each client’s data is sampled from a Dirichlet distribution $q \sim \operatorname{Dir}(\alpha p)$. Here, $p$ is the class prior, and $\alpha$ controls the degree of non-IID. Smaller $\alpha$ results in greater class imbalance, while larger $\alpha$ reduces the difference in data distribution among clients and more challenging local tasks. This approach effectively captures diverse and complex non-IID scenarios, making it a robust evaluation method. We study three different architectures: specifically VGG6 for CIFAR-10, ResNet-8 for CIFAR100, and ResNet-10 for TinyImageNet as in \citep{wu2023bold}.

\begin{table*}[!htbp]
\centering
\caption{Performance comparison on CIFAR-10, CIFAR-100, and TinyImageNet using different architectures (VGG6, ResNet-8, and ResNet-10, respectively), with 100 clients, each holding 100 samples, under varying values of $\alpha$.}
\label{tab:performance}
\resizebox{1.03\linewidth}{!}{
\begin{tabular}{l|ccc|ccc|ccc}
\toprule[1.5pt]
\multirow{2}{*}{Method} & \multicolumn{3}{c|}{CIFAR-10} & \multicolumn{3}{c|}{CIFAR-100} & \multicolumn{3}{c}{TinyImageNet} \\ \cline{2-10} 
 & $\alpha=0.1$ & $\alpha=0.5$ & $\alpha=1.0$ & $\alpha=0.01$ & $\alpha=0.1$ & $\alpha=0.5$ & $\alpha=0.01$ & $\alpha=0.1$ & $\alpha=0.5$ \\ \hline
Separate & 77.65$\pm$2.57 & 51.01$\pm$0.74 & 41.20$\pm$0.29 & 78.12$\pm$0.93 & 34.96$\pm$0.14 & 14.35$\pm$0.51 & 64.23$\pm$0.29 & 22.89$\pm$0.15 & 8.00$\pm$0.54 \\ 
FedAvg & 60.45$\pm$2.63 & 67.76$\pm$1.11 & 68.31$\pm$0.62 & 13.96$\pm$0.12 & 24.51$\pm$0.87 & 26.34$\pm$1.56 & 7.93$\pm$0.57 & 11.75$\pm$0.36 & 12.63$\pm$0.37 \\ 
FedRep & 81.09$\pm$1.94& 59.76$\pm$1.11& 51.47$\pm$0.42& 76.96$\pm$1.90& 34.31$\pm$0.16& 13.71$\pm$0.38& 60.24$\pm$1.53& 18.31$\pm$0.12& 8.09$\pm$0.15 \\ 
pFedSD & 80.82$\pm$1.87 &62.04$\pm$0.91 &60.60$\pm$0.65 & 73.22$\pm$1.11 & 38.62$\pm$0.36& 21.84$\pm$0.57 & 53.29$\pm$0.62& 28.12$\pm$0.43 & 16.23$\pm$0.35\\ 
FedBN & 85.83$\pm$1.28  & 72.69$\pm$0.75 & 68.66$\pm$0.56 & 79.28$\pm$0.72 & 43.08$\pm$0.45 & 24.45$\pm$0.61 & 72.55$\pm$0.63 & 36.83$\pm$0.22 & 20.73$\pm$1.23 \\ 
FedCAC & 84.74$\pm$1.46 & 70.79$\pm$0.61 & 66.93$\pm$0.65 & 78.87$\pm$0.50 & 43.96$\pm$0.54 & 25.31$\pm$0.54 & 68.93$\pm$0.56& 32.50$\pm$0.32 & 18.50$\pm$0.53 \\ \hline
LIPS & \textbf{86.21$\pm$1.66} & \textbf{74.78$\pm$0.68} & \textbf{72.07$\pm$0.38} & \textbf{81.39$\pm$0.43} & \textbf{47.84$\pm$0.47} & \textbf{28.97$\pm$1.02} & \textbf{74.83$\pm$0.08} & \textbf{40.61$\pm$0.54} & \textbf{23.53$\pm$0.15} \\

\bottomrule[1.5pt]
\end{tabular}
}
\end{table*}

We evaluate LIPS against several baselines, including FedAvg \citep{mcmahan2017communication}, FedRep \citep{collins2021exploiting}, FedBN \citep{lifedbn}, pFedSD \citep{jin2022personalized}, and FedCAC \citep{wu2023bold}. Additionally, we include a baseline, termed as ``Separate”, where models are trained locally on client data without any federated aggregation.
To highlight the effectiveness of client collaboration in the low-data regime, we assign a limited amount of data to each client in our main experiments. Specifically, for CIFAR-10 and CIFAR-100, each client is assigned 100 training samples, while for TinyImageNet, each client is assigned 300 training samples along with 400 test samples. Each task involves 100 clients, and the test data follows the same distribution as the training data to ensure consistent evaluation. Additionally, we extend our exploration to scenarios with 200 and 300 clients and varying training sample sizes of 60, 300, 500, and 700 per client.
We provide more details of the implementation in Appendix~\ref{app:impl_details}.

\subsection{Overall Performance}

In this section, we will thoughtfully evaluate the effectiveness of the proposed LIPS across CIFAR-10, CIFAR-100, and TinyImageNet datasets. Our evaluation considers a range of scenarios, including varying data distributions, different numbers of clients, and diverse numbers of data per client.

\subsubsection{Performance Across Datasets.} 
We evaluate our approach under Dirichlet non-IID scenarios with $\alpha$ values of {0.1, 0.5, 1.0} for CIFAR-10 and {0.01, 0.1, 0.5} for CIFAR-100 and TinyImageNet, as the same settings in \citep{wu2023bold}.
As shown in Table 2, most baselines achieve significant improvements over the ``Separate” approach in the Dirichlet non-IID scenario, particularly at higher $\alpha$ values, indicating that federated learning can effectively enhance performance.
However, the experimental results reveal notable differences in performance across datasets. For CIFAR-100 and TinyImageNet, the increased number of classes complicates local tasks and increases the risk of overfitting, which diminishes the effectiveness of methods such as FedRep and FedCAC. 

Additionally, under high $\alpha$, several of these methods perform worse than FedAvg, suggesting that they struggle to fully exploit shared knowledge when data is more uniformly distributed. In contrast, LIPS consistently achieves the best performance across these settings, highlighting its superior ability to support effective collaboration in low-data FL scenarios.

\subsubsection{Impact on Data Distribution.} 
As shown in Table~\ref{tab:performance}, LIPS consistently improves performance across various degrees of client data heterogeneity. It is particularly effective in more challenging scenarios involving highly diverse local data distributions, such as Dir($\alpha$=1.0) for CIFAR-10 and Dir($\alpha$=0.5) for CIFAR-100 and TinyImageNet. In these settings, LIPS achieves performance gains of 3–4\% over strong baselines.
By explicitly accounting for differences in client data and local tasks, LIPS enables more effective global aggregation and fosters collaboration even under significant distribution shifts. This leads to consistent performance gains across diverse federated learning scenarios.


\subsubsection{Impact on the Number of Clients.} 
To examine the impact of varying client numbers on the performance of LIPS, we conduct experiments on CIFAR-10 and CIFAR-100 under non-IID settings with Dir($\alpha$=0.5) and Dir($\alpha$=0.1), respectively. The number of clients is set to 100, 200 and 300, with each client having 100 training samples.
As shown in Table~\ref{tab:clients}, compared to FedBN, which is identical except for the absence of sparsity, the performance improvement achieved by LIPS becomes more pronounced as the number of clients increases. This is likely because a larger number of clients exacerbates the heterogeneity in data distributions, making it more challenging for traditional aggregation methods to converge effectively. LIPS, by dynamically introducing sparsity, redistributes learning capacity and fosters more meaningful updates, thereby mitigating the adverse effects of increased client diversity. These results highlight the scalability of LIPS and its effectiveness in various FL settings.

\begin{table*}[!t]
\centering
\caption{Comparison of LIPS and FedBN performance on CIFAR-10 with Dir($\alpha$=0.5) and CIFAR-100 with Dir($\alpha$=0.1), under different number of clients.}
\label{tab:clients}
\resizebox{\linewidth}{!}{
\begin{tabular}{l|ccc|ccc}
\toprule[1.5pt]
\multirow{2}{*}{Method} & \multicolumn{3}{c|}{CIFAR-10} & \multicolumn{3}{c}{CIFAR-100} \\ \cline{2-7} 
 & $N=100$ & $N=200$ & $N=300$ & $N=100$ & $N=200$ & $N=300$ \\ \hline
FedBN & 72.69 & 74.45 & 75.13 & 43.08 & 44.26 & 44.84 \\
LIPS & 74.78(\textcolor{blue}{2.09}$\uparrow$ ) & 77.56(\textcolor{blue}{3.11}$\uparrow$) & 78.61(\textcolor{blue}{3.48}$\uparrow$) & 47.84(\textcolor{blue}{4.76}$\uparrow$) & 49.58(\textcolor{blue}{5.32}$\uparrow$ ) & 50.36(\textcolor{blue}{5.52}$\uparrow$) \\ 
\bottomrule[1.5pt]
\end{tabular}
}
\vskip -0.2in
\end{table*}

\begin{wrapfigure}{r}{0.6\textwidth}
\vskip -0.2in
\begin{minipage}{0.6\textwidth}
    \centering
    \includegraphics[width=\textwidth]{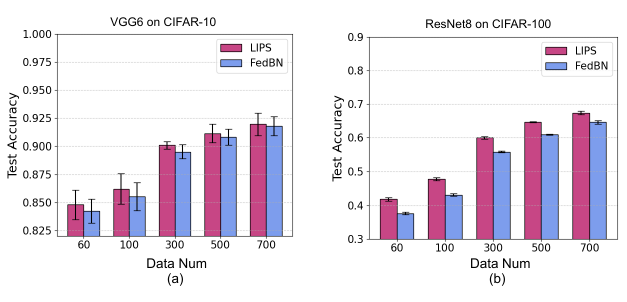}
    \vskip -0.2in
    \caption{Performance comparison across varying numbers of training samples (Data Num) per client (100, 300, 500, and 700 samples) on (a) CIFAR-10 and (b) CIFAR-100 datasets under a data distribution with Dir($\alpha$=0.1).}
    \label{fig: fd_data}
    \vskip -0.2in
\end{minipage}
\end{wrapfigure}

\subsubsection{Varying Numbers of Data per Client.}  
As discussed in Section~\ref{inertia_phenomenon}, smaller training datasets within clients exacerbate the layer-wise inertia phenomenon. To evaluate the effectiveness of the proposed LIPS method under different data regimes, we conduct experiments on CIFAR-10 and CIFAR-100 in non-IID settings with Dirichlet parameters of 0.1. Each client is assigned 60, 100, 300, 500, and 700 training samples, respectively.
The results, presented in Figure \ref{fig: fd_data}, demonstrate that LIPS achieves greater performance improvement under low-data settings compared with baseline. This can be attributed to the more pronounced layer-wise inertia phenomenon when client data is limited, which restricts the effectiveness of aggregation in FL. On the other hand, as the number of data increased per client, this effect diminishes—intermediate layers receive richer updates, alleviating stagnation (as shown in Figure \ref{fig:fd_datamodel}), and enabling more effective learning even with standard FL methods. Consequently, LIPS proves especially effective in low-data scenarios, where federated learning typically encounters greater challenges.

\subsection{Ablation Study}
\label{sec:ablation}

\begin{wraptable}{r}{0.6\textwidth}
\vskip -0.2in
\begin{minipage}{0.6\textwidth}
\centering
\caption{Performance comparison of various weight selection methods on CIFAR-10/100 with different values of $\alpha$.}
\label{tab:criteria}
\resizebox{\linewidth}{!}{
\begin{tabular}{l|cc|cc}
\toprule[1.5pt]
\multirow{2}{*}{Method} & \multicolumn{2}{c|}{CIFAR-10} & \multicolumn{2}{c}{CIFAR-100} \\ \cline{2-5} 
 & $\alpha=0.1$ & $\alpha=0.5$ & $\alpha=0.01$ & $\alpha=0.1$ \\ \hline
FedBN & 85.83$\pm$1.28  & 72.69$\pm$0.75 & 79.28$\pm$0.72 & 43.08$\pm$0.45 \\
$\tt LIPS_{Magnitude}$  & 86.01$\pm$1.18 & 72.56$\pm$0.49 & 79.40$\pm$0.91 & 43.31$\pm$0.04\\
$\tt LIPS_{Random}$  & 84.34$\pm$1.10 & 73.62$\pm$0.27 & 80.77$\pm$0.86 & \textbf{50.14}$\pm$0.47 \\ 
$\tt LIPS_{Sensitivity}$  & \textbf{86.21$\pm$1.66} & \textbf{74.78$\pm$0.68} & \textbf{81.39$\pm$0.43} & 47.84$\pm$0.47\\

\bottomrule[1.5pt]
\end{tabular}
}
\end{minipage}
\end{wraptable}

\textbf{Effect of weights selection methods.} 
We compare our sensitivity-based criterion with two alternative weight selection methods: Random, which randomly zeros out weights during sparsification; and Magnitude, which prunes weights based on their magnitude values. Experiments are conducted under non-IID settings with Dir($\alpha$=0.1) and Dir($\alpha$=0.5) on the CIFAR-10 and CIFAR-100 datasets, respectively.
As shown in Table \ref{tab:criteria}, the Magnitude method generally achieves accuracy comparable to the baseline (without sparsity). Interestingly, the Random method performs moderately well on CIFAR-100 but yields suboptimal results on CIFAR-10. In contrast, the Sensitivity-based method delivers consistently strong performance across both datasets and various non-IID settings. This indicates that sensitivity-based weight selection criterion effectively identifies critical parameters in federated learning.

\textbf{Effect of frequency $k$ and initial sparsity ratio $\tau_0$.}
We further investigate the impact of two key hyperparameters in LIPS: sparsification frequency $k$ and the initial sparsity ratio $\tau_0$, as detailed in Appendix~\ref{app:ablation}. We find that introduce sparsity every 5 communication rounds ($k=5$) strikes a good balance between model performance and adaptability across all datasets. This setting avoids overly frequent structural changes that may destabilize training while ensuring sufficient exploration of the parameter space. Regarding $\tau_0$, we adopt $\tau_0=0.5$ for CIFAR-10 and $\tau_0=0.7$ for CIFAR-100 and TinyImageNet, as these configurations consistently yield better performance. These results suggest that both the timing and degree of sparsification play important roles in enabling LIPS to effectively counteract overfitting and promote more meaningful global aggregation.

\begin{wraptable}{r}{0.6\textwidth}
\vskip -0.1in
\begin{minipage}{0.6\textwidth}
\centering
\caption{Performance comparison of weights initialization on CIFAR-10/100 with different values of $\alpha$.}
\label{tab:init}
\resizebox{\linewidth}{!}{
\begin{tabular}{l|cc|cc}
\toprule[1.5pt]
\multirow{2}{*}{Method} & \multicolumn{2}{c|}{CIFAR-10} & \multicolumn{2}{c}{CIFAR-100} \\ \cline{2-5} 
 & $\alpha=0.1$ & $\alpha=0.5$ & $\alpha=0.01$ & $\alpha=0.1$ \\ \hline
FedBN & 85.83$\pm$1.28  & 72.69$\pm$0.75 & 79.28$\pm$0.72 & 43.08$\pm$0.45 \\
$\tt LIPS_{w/ zero}$ & {86.21$\pm$1.66} & {74.78$\pm$0.68} & {81.39$\pm$0.43} & \textbf{47.84$\pm$0.47} \\
$\tt LIPS_{w/ init}$ & \textbf{87.10$\pm$1.38} & \textbf{75.31$\pm$0.67} & \textbf{81.45$\pm$0.82} & 47.80$\pm$0.62 \\

\bottomrule[1.5pt]
\end{tabular}
}
\end{minipage}
\vskip -0.1in
\end{wraptable}

\textbf{Effect of weights initialization.} 
When introducing sparsity in federated learning, an important design choice is how to initialize the weights that are zeroed out during sparsification for subsequent local training. We compare two strategies: (1) initializing these weights to zero ($\tt LIPS_{w/ zero}$) and (2) restoring them to their original initialization values used before sparsification ($\tt LIPS_{w/ init}$). As shown in Table~\ref{tab:init}, in general, $\tt LIPS_{w/ init}$ achieves slightly better performance than zero initialization in most cases.
However, $\tt LIPS_{w/ init}$ requires additional storage to retain the original initialization, which increases memory overhead. Considering the trade-off between efficiency and performance, we adopt zero initialization as the default setting in all experiments. That said, using the original initialization for reset weights presents a promising avenue for future research aimed at enhancing model performance.
\begin{figure*}[!htb]
\vskip -0.1in
    \centering
    \includegraphics[width=\textwidth]{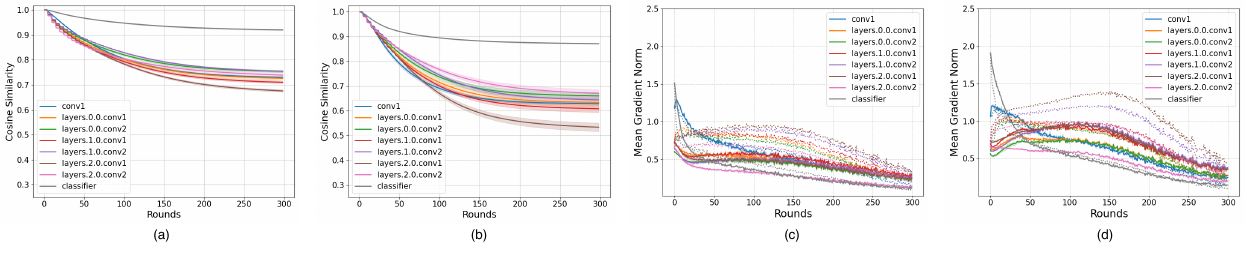}
    \vskip -0.15in
    \caption{Layer-wise cosine similarity throughout the training process for LIPS on CIFAR-100 datasets under Dir($\alpha$=0.01) and Dir($\alpha$=0.1), shown in (a) and (b), respectively. A comparison of layer-wise mean gradient norms across all clients between FedBN (solid line) and LIPS (dotted line) after each training round under Dir($\alpha$=0.01) and Dir($\alpha$=0.1) in (c) and (d), respectively.}
    \label{fig: fd_visual}
    \vskip -0.15in
\end{figure*}

\subsection{Visualizing the Impact of LIPS on Layer Dynamics}

\textbf{Layer-wise cosine similarity.} For LIPS, we track the global model’s layer-wise cosine similarity relative to the 2nd round states across the training process on CIFAR-100 datasets, using ResNet-8 architectures under Dir($\alpha$=0.01) and Dir($\alpha$=0.1). The experiments are conducted following the same configuration as in Figure \ref{fig: fd_inertia}. 
As illustrated in Figure~\ref{fig: fd_visual} (b), introducing sparsity during training significantly reduces the cosine similarity in the middle layers compared to their counterparts in Figure~\ref{fig: fd_inertia} (b). This reduction suggests that sparsity effectively mitigates the Layer-wise Inertia phenomenon by facilitating more substantial updates in these layers, thereby enhancing collaboration during global aggregation.

\textbf{Layer-wise mean gradient norm.} Under the same experimental settings, we further analyze the layer-wise mean gradient norm across all clients after each round of training, comparing FedBN and LIPS in Figure \ref{fig: fd_visual} (c) and (d). Our results reveal that, in contrast to FedBN, which does not utilize sparsity, LIPS increases the gradient norm during local client training. This increase in gradient norm is a crucial factor for activating weight updates, particularly for layers that typically stagnate under traditional FL settings. By fostering more active weight updates, LIPS improves the overall effectiveness of aggregation in federated learning, leading to a more robust global model that performs better across diverse client scenarios.

Additional results on other datasets and architectures are provided in Appendix \ref{app:visual}.



\section{Related Work}
\label{app:related workd}

\subsection{Personalization in Federated Learning}

Personalized federated learning aims to address the issue of data heterogeneity by adapting clients to their local data distribution \citep{tan2022towards}. Common approaches include multitask learning \citep{smith2017federated, agarwal2020federated}, clustering \citep{duan2021flexible, ghosh2020efficient, mansour2020three}, transfer learning \citep{yu2020salvaging, zhang2021parameterized}, meta learning \citep{singhal2021federated, jiang2019improving}, etc. Recently, partial model personalization has gained attention for improving client model performance by adapting specific components of the model to local tasks. For instance, approaches like FedRep \citep{collins2021exploiting} and FedPer \citep{arivazhagan2019federated} focus on personalizing certain layers, such as classifier layers, while preserving globally shared representations. Furthermore, advanced aggregation methods, such as FedProx \citep{li2020federated} and Scaffold \citep{karimireddy2020scaffold}, incorporate regularization terms or variance reduction techniques to balance local and global learning objectives effectively. While these approaches tackle specific aspects of model adaptation and aggregation, they often overlook the layer-wise dynamics that critically influence aggregation effectiveness.

\subsection{Federated Learning with Limited Data }

Federated learning in low-data settings presents a unique set of challenges, where individual clients possess only a small number of samples. This amplifies the risks of overfitting and limits the representativeness of local updates, undermining the benefits of collaboration. Prior works such as FedMix \citep{yoon2021fedmix}, FedGEN \citep{venkateswaran2023fedgen}, and FedNTD \citep{lee2022preservation} explore data augmentation and generative modeling to alleviate data scarcity. Other efforts focus on regularization-based methods \citep{li2020federated, jeon2023federated, yuan2022what, li2021ditto} to improve model generalization under limited supervision. Despite these advances, few works systematically examine how limited data influences layer-wise training dynamics during federated optimization. Our work addresses this gap by identifying the Layer-wise Inertia Phenomenon in low-data FL and proposing a sparsity-driven solution to enhance model adaptation and aggregation efficacy in such constrained environments.

\subsection{Sparsity in Federated Learning}

Sparsity has been extensively explored in FL, primarily as a strategy to enhance communication efficiency by reducing the amount of information exchanged between clients and the server. For instance, gradient sparsification techniques selectively transmit only the most significant gradients \citep{mitra2021linear, han2020adaptive}, while model pruning methods reduce the model size by eliminating redundant weights during communication rounds \citep{babakniya2023revisiting, li2021lotteryfl, jiang2022model, chen2023efficient}. Additionally, recent advancements extend the benefits of sparsity beyond communication efficiency to improve local training efficiency \citep{bibikar2022federated, huang2022achieving, dai2022dispfl, kuo2024federated} through sparse training methods \citep{mocanu2018scalable, evci2020rigging, xiao2022dynamic, LiuCCCXWKPMW23, wu2025dynamic}. While these approaches effectively address bandwidth constraints and computational costs, their primary focus on optimizing resource usage.
In contrast, our work aims to enhance aggregation in FL by dynamically introducing transient sparsity during training. This approach fosters more meaningful updates across layers, enabling more effective aggregation and improving the performance of the global model in heterogeneous federated learning scenarios.

\section{Conclusion}

In this work, we uncovered the Layer-wise Inertia Phenomenon in low-data federated learning (FL), where middle layers of the global model exhibit stagnation after early training rounds, severely impairing the exchange of meaningful updates across clients.
To address this issue, we proposed a simple yet effective method, Layer-wise Inertia with Sparsity (LIPS), which periodically introduces transient sparsity during training to stimulate meaningful updates and mitigate the layer-wise inertia. Through extensive experiments, we showed that LIPS improves model performance across diverse non-IID settings by activating underutilized layers and enhancing the quality of aggregation. 
We also include a discussion of future directions and limitations in Appendix~\ref{app:limits}.

\section*{Acknowledgements}
This work is part of the research program ‘MegaMind - Measuring, Gathering, Mining and Integrating Data for Self-management in the Edge of the Electricity System’, (partly) financed by the Dutch Research Council (NWO) through the Perspectief program under number P19-25. This work is partly supported by the SmartCHANGE project, funded within EU’s Horizon Europe research program (GA No. 101080965).
Elena Mocanu is partly supported by the Modular Integrated Sustainable Datacenter (MISD) project funded by the Dutch Ministry of Economic Affairs and Climate under the European Important Projects of Common European Interest - Cloud Infrastructure and Services (IPCEI-CIS) program for 2024-2029.
This work used the Dutch national e-infrastructure with the support of the SURF Cooperative, using grant no. EINF-12291.

{
\bibliography{example_paper}

\begin{thebibliography}{70}
\providecommand{\natexlab}[1]{#1}
\providecommand{\url}[1]{\texttt{#1}}
\expandafter\ifx\csname urlstyle\endcsname\relax
  \providecommand{\doi}[1]{doi: #1}\else
  \providecommand{\doi}{doi: \begingroup \urlstyle{rm}\Url}\fi

\bibitem[Agarwal et~al.(2020)Agarwal, Langford, and Wei]{agarwal2020federated}
Alekh Agarwal, John Langford, and Chen-Yu Wei.
\newblock Federated residual learning.
\newblock \emph{arXiv preprint arXiv:2003.12880}, 2020.

\bibitem[Arivazhagan et~al.(2019)Arivazhagan, Aggarwal, Singh, and Choudhary]{arivazhagan2019federated}
Manoj~Ghuhan Arivazhagan, Vinay Aggarwal, Aaditya~Kumar Singh, and Sunav Choudhary.
\newblock Federated learning with personalization layers.
\newblock \emph{arXiv preprint arXiv:1912.00818}, 2019.

\bibitem[Babakniya et~al.(2023)Babakniya, Kundu, Prakash, Niu, and Avestimehr]{babakniya2023revisiting}
Sara Babakniya, Souvik Kundu, Saurav Prakash, Yue Niu, and Salman Avestimehr.
\newblock Revisiting sparsity hunting in federated learning: Why does sparsity consensus matter?
\newblock \emph{Transactions on Machine Learning Research}, 2023.

\bibitem[Bibikar et~al.(2022)Bibikar, Vikalo, Wang, and Chen]{bibikar2022federated}
Sameer Bibikar, Haris Vikalo, Zhangyang Wang, and Xiaohan Chen.
\newblock Federated dynamic sparse training: Computing less, communicating less, yet learning better.
\newblock In \emph{Proceedings of the AAAI Conference on Artificial Intelligence}, volume~36, pages 6080--6088, 2022.

\bibitem[Chen et~al.(2023)Chen, Yao, Gao, Ding, and Li]{chen2023efficient}
Daoyuan Chen, Liuyi Yao, Dawei Gao, Bolin Ding, and Yaliang Li.
\newblock Efficient personalized federated learning via sparse model-adaptation.
\newblock In \emph{International Conference on Machine Learning}, pages 5234--5256. PMLR, 2023.

\bibitem[Chen et~al.(2022)Chen, Zhang, pengjun wang, Balachandra, Ma, Wang, and Wang]{chen2022sparsity}
Tianlong Chen, Zhenyu Zhang, pengjun wang, Santosh Balachandra, Haoyu Ma, Zehao Wang, and Zhangyang Wang.
\newblock Sparsity winning twice: Better robust generalization from more efficient training.
\newblock In \emph{International Conference on Learning Representations}, 2022.
\newblock URL \url{https://openreview.net/forum?id=SYuJXrXq8tw}.

\bibitem[Chen et~al.(2025)Chen, Hu, Zhang, Wang, Li, and Chen]{chen2025streamlining}
Xiaodong Chen, Yuxuan Hu, Jing Zhang, Yanling Wang, Cuiping Li, and Hong Chen.
\newblock Streamlining redundant layers to compress large language models.
\newblock In \emph{The Thirteenth International Conference on Learning Representations}, 2025.
\newblock URL \url{https://openreview.net/forum?id=IC5RJvRoMp}.

\bibitem[Collins et~al.(2021)Collins, Hassani, Mokhtari, and Shakkottai]{collins2021exploiting}
Liam Collins, Hamed Hassani, Aryan Mokhtari, and Sanjay Shakkottai.
\newblock Exploiting shared representations for personalized federated learning.
\newblock In \emph{International conference on machine learning}, pages 2089--2099. PMLR, 2021.

\bibitem[Dai et~al.(2022)Dai, Shen, He, Tian, and Tao]{dai2022dispfl}
Rong Dai, Li~Shen, Fengxiang He, Xinmei Tian, and Dacheng Tao.
\newblock Dispfl: Towards communication-efficient personalized federated learning via decentralized sparse training.
\newblock In \emph{International Conference on Machine Learning}, pages 4587--4604. PMLR, 2022.

\bibitem[Duan et~al.(2021)Duan, Liu, Ji, Wu, Liang, Chen, Tan, and Ren]{duan2021flexible}
Moming Duan, Duo Liu, Xinyuan Ji, Yu~Wu, Liang Liang, Xianzhang Chen, Yujuan Tan, and Ao~Ren.
\newblock Flexible clustered federated learning for client-level data distribution shift.
\newblock \emph{IEEE Transactions on Parallel and Distributed Systems}, 33\penalty0 (11):\penalty0 2661--2674, 2021.

\bibitem[Evci et~al.(2020)Evci, Gale, Menick, Castro, and Elsen]{evci2020rigging}
Utku Evci, Trevor Gale, Jacob Menick, Pablo~Samuel Castro, and Erich Elsen.
\newblock Rigging the lottery: Making all tickets winners.
\newblock In \emph{International conference on machine learning}, pages 2943--2952. PMLR, 2020.

\bibitem[Evci et~al.(2022)Evci, Ioannou, Keskin, and Dauphin]{evci2022gradient}
Utku Evci, Yani Ioannou, Cem Keskin, and Yann Dauphin.
\newblock Gradient flow in sparse neural networks and how lottery tickets win.
\newblock In \emph{Proceedings of the AAAI conference on artificial intelligence}, volume~36, pages 6577--6586, 2022.

\bibitem[Ghosh et~al.(2020)Ghosh, Chung, Yin, and Ramchandran]{ghosh2020efficient}
Avishek Ghosh, Jichan Chung, Dong Yin, and Kannan Ramchandran.
\newblock An efficient framework for clustered federated learning.
\newblock \emph{Advances in Neural Information Processing Systems}, 33:\penalty0 19586--19597, 2020.

\bibitem[Gromov et~al.(2025)Gromov, Tirumala, Shapourian, Glorioso, and Roberts]{gromov2025the}
Andrey Gromov, Kushal Tirumala, Hassan Shapourian, Paolo Glorioso, and Dan Roberts.
\newblock The unreasonable ineffectiveness of the deeper layers.
\newblock In \emph{The Thirteenth International Conference on Learning Representations}, 2025.
\newblock URL \url{https://openreview.net/forum?id=ngmEcEer8a}.

\bibitem[Han et~al.(2020)Han, Wang, and Leung]{han2020adaptive}
Pengchao Han, Shiqiang Wang, and Kin~K Leung.
\newblock Adaptive gradient sparsification for efficient federated learning: An online learning approach.
\newblock In \emph{2020 IEEE 40th international conference on distributed computing systems (ICDCS)}, pages 300--310. IEEE, 2020.

\bibitem[He et~al.(2016)He, Zhang, Ren, and Sun]{he2016deep}
Kaiming He, Xiangyu Zhang, Shaoqing Ren, and Jian Sun.
\newblock Deep residual learning for image recognition.
\newblock In \emph{Proceedings of the IEEE conference on computer vision and pattern recognition}, pages 770--778, 2016.

\bibitem[Hsu et~al.(2019)Hsu, Qi, and Brown]{hsu2019measuring}
Tzu-Ming~Harry Hsu, Hang Qi, and Matthew Brown.
\newblock Measuring the effects of non-identical data distribution for federated visual classification.
\newblock \emph{arXiv preprint arXiv:1909.06335}, 2019.

\bibitem[Huang et~al.(2022)Huang, Liu, Shen, He, Lin, and Tao]{huang2022achieving}
Tiansheng Huang, Shiwei Liu, Li~Shen, Fengxiang He, Weiwei Lin, and Dacheng Tao.
\newblock Achieving personalized federated learning with sparse local models.
\newblock \emph{arXiv preprint arXiv:2201.11380}, 2022.

\bibitem[Jeon et~al.(2023)Jeon, Hong, Yun, and Kim]{jeon2023federated}
Insu Jeon, Minui Hong, Junhyeog Yun, and Gunhee Kim.
\newblock Federated learning via meta-variational dropout.
\newblock In \emph{Thirty-seventh Conference on Neural Information Processing Systems}, 2023.
\newblock URL \url{https://openreview.net/forum?id=VNyKBipt91}.

\bibitem[Jiang et~al.(2025)Jiang, Zhou, and Zhu]{jiang2025tracing}
Jiachen Jiang, Jinxin Zhou, and Zhihui Zhu.
\newblock Tracing representation progression: Analyzing and enhancing layer-wise similarity.
\newblock In \emph{The Thirteenth International Conference on Learning Representations}, 2025.
\newblock URL \url{https://openreview.net/forum?id=vVxeFSR4fU}.

\bibitem[Jiang and Borcea(2023)]{jiang2023complement}
Xiaopeng Jiang and Cristian Borcea.
\newblock Complement sparsification: Low-overhead model pruning for federated learning.
\newblock In \emph{Proceedings of the AAAI Conference on Artificial Intelligence}, volume~37, pages 8087--8095, 2023.

\bibitem[Jiang et~al.(2019)Jiang, Kone{\v{c}}n{\`y}, Rush, and Kannan]{jiang2019improving}
Yihan Jiang, Jakub Kone{\v{c}}n{\`y}, Keith Rush, and Sreeram Kannan.
\newblock Improving federated learning personalization via model agnostic meta learning.
\newblock \emph{arXiv preprint arXiv:1909.12488}, 2019.

\bibitem[Jiang et~al.(2022)Jiang, Wang, Valls, Ko, Lee, Leung, and Tassiulas]{jiang2022model}
Yuang Jiang, Shiqiang Wang, Victor Valls, Bong~Jun Ko, Wei-Han Lee, Kin~K Leung, and Leandros Tassiulas.
\newblock Model pruning enables efficient federated learning on edge devices.
\newblock \emph{IEEE Transactions on Neural Networks and Learning Systems}, 34\penalty0 (12):\penalty0 10374--10386, 2022.

\bibitem[Jin et~al.(2022)Jin, Bai, Yao, Dai, Gu, Yu, and Sun]{jin2022personalized}
Hai Jin, Dongshan Bai, Dezhong Yao, Yutong Dai, Lin Gu, Chen Yu, and Lichao Sun.
\newblock Personalized edge intelligence via federated self-knowledge distillation.
\newblock \emph{IEEE Transactions on Parallel and Distributed Systems}, 34\penalty0 (2):\penalty0 567--580, 2022.

\bibitem[Karimireddy et~al.(2020)Karimireddy, Kale, Mohri, Reddi, Stich, and Suresh]{karimireddy2020scaffold}
Sai~Praneeth Karimireddy, Satyen Kale, Mehryar Mohri, Sashank Reddi, Sebastian Stich, and Ananda~Theertha Suresh.
\newblock Scaffold: Stochastic controlled averaging for federated learning.
\newblock In \emph{International conference on machine learning}, pages 5132--5143. PMLR, 2020.

\bibitem[Krizhevsky et~al.(2009)Krizhevsky, Hinton, et~al.]{krizhevsky2009learning}
Alex Krizhevsky, Geoffrey Hinton, et~al.
\newblock Learning multiple layers of features from tiny images.
\newblock 2009.

\bibitem[Krizhevsky et~al.(2010)Krizhevsky, Nair, and Hinton]{krizhevsky2010cifar}
Alex Krizhevsky, Vinod Nair, and Geoffrey Hinton.
\newblock Cifar-10 (canadian institute for advanced research).
\newblock \emph{URL http://www. cs. toronto. edu/kriz/cifar. html}, 5\penalty0 (4):\penalty0 1, 2010.

\bibitem[Kuo et~al.(2024)Kuo, Raje, Rajesh, and Smith]{kuo2024federated}
Kevin Kuo, Arian Raje, Kousik Rajesh, and Virginia Smith.
\newblock Federated lora with sparse communication.
\newblock \emph{arXiv preprint arXiv:2406.05233}, 2024.

\bibitem[Le and Yang(2015)]{le2015tiny}
Yann Le and Xuan Yang.
\newblock Tiny imagenet visual recognition challenge.
\newblock \emph{CS 231N}, 7\penalty0 (7):\penalty0 3, 2015.

\bibitem[Lee et~al.(2022)Lee, Jeong, Shin, Bae, and Yun]{lee2022preservation}
Gihun Lee, Minchan Jeong, Yongjin Shin, Sangmin Bae, and Se-Young Yun.
\newblock Preservation of the global knowledge by not-true distillation in federated learning.
\newblock In Alice~H. Oh, Alekh Agarwal, Danielle Belgrave, and Kyunghyun Cho, editors, \emph{Advances in Neural Information Processing Systems}, 2022.
\newblock URL \url{https://openreview.net/forum?id=qw3MZb1Juo}.

\bibitem[Lee et~al.(2019)Lee, Ajanthan, and Torr]{lee2019snip}
N~Lee, T~Ajanthan, and P~Torr.
\newblock Snip: single-shot network pruning based on connection sensitivity.
\newblock In \emph{International Conference on Learning Representations}. Open Review, 2019.

\bibitem[Li et~al.(2021{\natexlab{a}})Li, Sun, Wang, Duan, Li, Chen, and Li]{li2021lotteryfl}
Ang Li, Jingwei Sun, Binghui Wang, Lin Duan, Sicheng Li, Yiran Chen, and Hai Li.
\newblock Lotteryfl: Empower edge intelligence with personalized and communication-efficient federated learning.
\newblock In \emph{2021 IEEE/ACM Symposium on Edge Computing (SEC)}, pages 68--79. IEEE, 2021{\natexlab{a}}.

\bibitem[Li et~al.(2022)Li, Diao, Chen, and He]{li2022federated}
Qinbin Li, Yiqun Diao, Quan Chen, and Bingsheng He.
\newblock Federated learning on non-iid data silos: An experimental study.
\newblock In \emph{2022 IEEE 38th international conference on data engineering (ICDE)}, pages 965--978. IEEE, 2022.

\bibitem[Li et~al.(2020)Li, Sahu, Zaheer, Sanjabi, Talwalkar, and Smith]{li2020federated}
Tian Li, Anit~Kumar Sahu, Manzil Zaheer, Maziar Sanjabi, Ameet Talwalkar, and Virginia Smith.
\newblock Federated optimization in heterogeneous networks.
\newblock \emph{Proceedings of Machine learning and systems}, 2:\penalty0 429--450, 2020.

\bibitem[Li et~al.(2021{\natexlab{b}})Li, Hu, Beirami, and Smith]{li2021ditto}
Tian Li, Shengyuan Hu, Ahmad Beirami, and Virginia Smith.
\newblock Ditto: Fair and robust federated learning through personalization.
\newblock In \emph{International conference on machine learning}, pages 6357--6368. PMLR, 2021{\natexlab{b}}.

\bibitem[Li et~al.(2021{\natexlab{c}})Li, JIANG, Zhang, Kamp, and Dou]{lifedbn}
Xiaoxiao Li, Meirui JIANG, Xiaofei Zhang, Michael Kamp, and Qi~Dou.
\newblock Fedbn: Federated learning on non-iid features via local batch normalization.
\newblock In \emph{International Conference on Learning Representations}, 2021{\natexlab{c}}.

\bibitem[Lin et~al.(2020)Lin, Kong, Stich, and Jaggi]{lin2020ensemble}
Tao Lin, Lingjing Kong, Sebastian~U Stich, and Martin Jaggi.
\newblock Ensemble distillation for robust model fusion in federated learning.
\newblock \emph{Advances in neural information processing systems}, 33:\penalty0 2351--2363, 2020.

\bibitem[Liu et~al.(2023)Liu, Chen, Chen, Chen, Xiao, Wu, Kärkkäinen, Pechenizkiy, Mocanu, and Wang]{LiuCCCXWKPMW23}
Shiwei Liu, Tianlong Chen, Xiaohan Chen, Xuxi Chen, Qiao Xiao, Boqian Wu, Tommi Kärkkäinen, Mykola Pechenizkiy, Decebal~Constantin Mocanu, and Zhangyang Wang.
\newblock More convnets in the 2020s: Scaling up kernels beyond 51x51 using sparsity.
\newblock In \emph{ICLR}, 2023.
\newblock URL \url{https://openreview.net/pdf?id=bXNl-myZkJl}.

\bibitem[Long et~al.(2020)Long, Tan, Jiang, and Zhang]{long2020federated}
Guodong Long, Yue Tan, Jing Jiang, and Chengqi Zhang.
\newblock Federated learning for open banking.
\newblock In \emph{Federated learning: privacy and incentive}, pages 240--254. Springer, 2020.

\bibitem[Lu et~al.(2024)Lu, Jiang, Mao, Tang, Chen, Cui, and Wang]{lu2024data}
Rongwei Lu, Yutong Jiang, Yinan Mao, Chen Tang, Bin Chen, Laizhong Cui, and Zhi Wang.
\newblock Data-aware gradient compression for fl in communication-constrained mobile computing.
\newblock \emph{IEEE Transactions on Mobile Computing}, 2024.

\bibitem[Ma et~al.(2022)Ma, Zhang, Guo, and Xu]{ma2022layer}
Xiaosong Ma, Jie Zhang, Song Guo, and Wenchao Xu.
\newblock Layer-wised model aggregation for personalized federated learning.
\newblock In \emph{Proceedings of the IEEE/CVF conference on computer vision and pattern recognition}, pages 10092--10101, 2022.

\bibitem[Mansour et~al.(2020)Mansour, Mohri, Ro, and Suresh]{mansour2020three}
Yishay Mansour, Mehryar Mohri, Jae Ro, and Ananda~Theertha Suresh.
\newblock Three approaches for personalization with applications to federated learning.
\newblock \emph{arXiv preprint arXiv:2002.10619}, 2020.

\bibitem[McMahan et~al.(2017)McMahan, Moore, Ramage, Hampson, and y~Arcas]{mcmahan2017communication}
Brendan McMahan, Eider Moore, Daniel Ramage, Seth Hampson, and Blaise~Aguera y~Arcas.
\newblock Communication-efficient learning of deep networks from decentralized data.
\newblock In \emph{Artificial intelligence and statistics}, pages 1273--1282. PMLR, 2017.

\bibitem[McMahan et~al.(2016)McMahan, Moore, Ramage, and y~Arcas]{mcmahan2016federated}
H~Brendan McMahan, Eider Moore, Daniel Ramage, and Blaise~Ag{\"u}era y~Arcas.
\newblock Federated learning of deep networks using model averaging.
\newblock \emph{arXiv preprint arXiv:1602.05629}, 2\penalty0 (2), 2016.

\bibitem[Min and Wang(2025)]{min2025docs}
Zeping Min and Xinshang Wang.
\newblock {DOCS}: Quantifying weight similarity for deeper insights into large language models.
\newblock In \emph{The Thirteenth International Conference on Learning Representations}, 2025.
\newblock URL \url{https://openreview.net/forum?id=XBHoaHlGQM}.

\bibitem[Mitra et~al.(2021)Mitra, Jaafar, Pappas, and Hassani]{mitra2021linear}
Aritra Mitra, Rayana Jaafar, George~J Pappas, and Hamed Hassani.
\newblock Linear convergence in federated learning: Tackling client heterogeneity and sparse gradients.
\newblock \emph{Advances in Neural Information Processing Systems}, 34:\penalty0 14606--14619, 2021.

\bibitem[Mocanu et~al.(2018)Mocanu, Mocanu, Stone, Nguyen, Gibescu, and Liotta]{mocanu2018scalable}
Decebal~Constantin Mocanu, Elena Mocanu, Peter Stone, Phuong~H Nguyen, Madeleine Gibescu, and Antonio Liotta.
\newblock Scalable training of artificial neural networks with adaptive sparse connectivity inspired by network science.
\newblock \emph{Nature communications}, 9\penalty0 (1):\penalty0 2383, 2018.

\bibitem[Nguyen et~al.(2022)Nguyen, Pham, Pathirana, Ding, Seneviratne, Lin, Dobre, and Hwang]{nguyen2022federated}
Dinh~C Nguyen, Quoc-Viet Pham, Pubudu~N Pathirana, Ming Ding, Aruna Seneviratne, Zihuai Lin, Octavia Dobre, and Won-Joo Hwang.
\newblock Federated learning for smart healthcare: A survey.
\newblock \emph{ACM Computing Surveys (Csur)}, 55\penalty0 (3):\penalty0 1--37, 2022.

\bibitem[Nowak et~al.(2024)Nowak, Grooten, Mocanu, and Tabor]{nowak2024fantastic}
Aleksandra Nowak, Bram Grooten, Decebal~Constantin Mocanu, and Jacek Tabor.
\newblock Fantastic weights and how to find them: Where to prune in dynamic sparse training.
\newblock \emph{Advances in Neural Information Processing Systems}, 36, 2024.

\bibitem[Rieke et~al.(2020)Rieke, Hancox, Li, Milletari, Roth, Albarqouni, Bakas, Galtier, Landman, Maier-Hein, et~al.]{rieke2020future}
Nicola Rieke, Jonny Hancox, Wenqi Li, Fausto Milletari, Holger~R Roth, Shadi Albarqouni, Spyridon Bakas, Mathieu~N Galtier, Bennett~A Landman, Klaus Maier-Hein, et~al.
\newblock The future of digital health with federated learning.
\newblock \emph{NPJ digital medicine}, 3\penalty0 (1):\penalty0 1--7, 2020.

\bibitem[Sadilek et~al.(2021)Sadilek, Liu, Nguyen, Kamruzzaman, Serghiou, Rader, Ingerman, Mellem, Kairouz, Nsoesie, et~al.]{sadilek2021privacy}
Adam Sadilek, Luyang Liu, Dung Nguyen, Methun Kamruzzaman, Stylianos Serghiou, Benjamin Rader, Alex Ingerman, Stefan Mellem, Peter Kairouz, Elaine~O Nsoesie, et~al.
\newblock Privacy-first health research with federated learning.
\newblock \emph{NPJ digital medicine}, 4\penalty0 (1):\penalty0 132, 2021.

\bibitem[Simonyan and Zisserman(2015)]{simonyan2014very}
Karen Simonyan and Andrew Zisserman.
\newblock Very deep convolutional networks for large-scale image recognition.
\newblock In Yoshua Bengio and Yann LeCun, editors, \emph{3rd International Conference on Learning Representations, {ICLR}}, 2015.

\bibitem[Singhal et~al.(2021)Singhal, Sidahmed, Garrett, Wu, Rush, and Prakash]{singhal2021federated}
Karan Singhal, Hakim Sidahmed, Zachary Garrett, Shanshan Wu, John Rush, and Sushant Prakash.
\newblock Federated reconstruction: Partially local federated learning.
\newblock \emph{Advances in Neural Information Processing Systems}, 34:\penalty0 11220--11232, 2021.

\bibitem[Smith et~al.(2017)Smith, Chiang, Sanjabi, and Talwalkar]{smith2017federated}
Virginia Smith, Chao-Kai Chiang, Maziar Sanjabi, and Ameet~S Talwalkar.
\newblock Federated multi-task learning.
\newblock \emph{Advances in neural information processing systems}, 30, 2017.

\bibitem[Tamirisa et~al.(2024)Tamirisa, Xie, Bao, Zhou, Arel, and Shamsian]{tamirisa2024fedselect}
Rishub Tamirisa, Chulin Xie, Wenxuan Bao, Andy Zhou, Ron Arel, and Aviv Shamsian.
\newblock Fedselect: Personalized federated learning with customized selection of parameters for fine-tuning.
\newblock In \emph{Proceedings of the IEEE/CVF Conference on Computer Vision and Pattern Recognition}, pages 23985--23994, 2024.

\bibitem[Tan et~al.(2022)Tan, Yu, Cui, and Yang]{tan2022towards}
Alysa~Ziying Tan, Han Yu, Lizhen Cui, and Qiang Yang.
\newblock Towards personalized federated learning.
\newblock \emph{IEEE transactions on neural networks and learning systems}, 34\penalty0 (12):\penalty0 9587--9603, 2022.

\bibitem[Venkateswaran et~al.(2023)Venkateswaran, Isahagian, Muthusamy, and Venkatasubramanian]{venkateswaran2023fedgen}
Praveen Venkateswaran, Vatche Isahagian, Vinod Muthusamy, and Nalini Venkatasubramanian.
\newblock Fedgen: Generalizable federated learning for sequential data.
\newblock In \emph{2023 IEEE 16th International Conference on Cloud Computing (CLOUD)}, pages 308--318. IEEE, 2023.

\bibitem[Wangni et~al.(2018)Wangni, Wang, Liu, and Zhang]{wangni2018gradient}
Jianqiao Wangni, Jialei Wang, Ji~Liu, and Tong Zhang.
\newblock Gradient sparsification for communication-efficient distributed optimization.
\newblock \emph{Advances in Neural Information Processing Systems}, 31, 2018.

\bibitem[Wen et~al.(2022)Wen, Jeon, and Huang]{wen2022federated}
Dingzhu Wen, Ki-Jun Jeon, and Kaibin Huang.
\newblock Federated dropout—a simple approach for enabling federated learning on resource constrained devices.
\newblock \emph{IEEE wireless communications letters}, 11\penalty0 (5):\penalty0 923--927, 2022.

\bibitem[Wen et~al.(2023)Wen, Zhang, Lan, Cui, Cai, and Zhang]{wen2023survey}
Jie Wen, Zhixia Zhang, Yang Lan, Zhihua Cui, Jianghui Cai, and Wensheng Zhang.
\newblock A survey on federated learning: challenges and applications.
\newblock \emph{International Journal of Machine Learning and Cybernetics}, 14\penalty0 (2):\penalty0 513--535, 2023.

\bibitem[Wu et~al.(2025)Wu, Xiao, Wang, Strisciuglio, Pechenizkiy, van Keulen, Mocanu, and Mocanu]{wu2025dynamic}
Boqian Wu, Qiao Xiao, Shunxin Wang, Nicola Strisciuglio, Mykola Pechenizkiy, Maurice van Keulen, Decebal~Constantin Mocanu, and Elena Mocanu.
\newblock Dynamic sparse training versus dense training: The unexpected winner in image corruption robustness.
\newblock In \emph{The Thirteenth International Conference on Learning Representations}, 2025.
\newblock URL \url{https://openreview.net/forum?id=daUQ7vmGap}.

\bibitem[Wu et~al.(2023)Wu, Liu, Niu, Zhu, and Tang]{wu2023bold}
Xinghao Wu, Xuefeng Liu, Jianwei Niu, Guogang Zhu, and Shaojie Tang.
\newblock Bold but cautious: Unlocking the potential of personalized federated learning through cautiously aggressive collaboration.
\newblock In \emph{Proceedings of the IEEE/CVF International Conference on Computer Vision}, pages 19375--19384, 2023.

\bibitem[Xiao et~al.(2022)Xiao, Wu, Zhang, Liu, Pechenizkiy, Mocanu, and Mocanu]{xiao2022dynamic}
Qiao Xiao, Boqian Wu, Yu~Zhang, Shiwei Liu, Mykola Pechenizkiy, Elena Mocanu, and Decebal~Constantin Mocanu.
\newblock Dynamic sparse network for time series classification: Learning what to {\textquotedblleft}see{\textquotedblright}.
\newblock In Alice~H. Oh, Alekh Agarwal, Danielle Belgrave, and Kyunghyun Cho, editors, \emph{Advances in Neural Information Processing Systems}, 2022.
\newblock URL \url{https://openreview.net/forum?id=ZxOO5jfqSYw}.

\bibitem[Yang et~al.(2019)Yang, Liu, Chen, and Tong]{yang2019federated}
Qiang Yang, Yang Liu, Tianjian Chen, and Yongxin Tong.
\newblock Federated machine learning: Concept and applications.
\newblock \emph{ACM Transactions on Intelligent Systems and Technology (TIST)}, 10\penalty0 (2):\penalty0 1--19, 2019.

\bibitem[Yoon et~al.(2021)Yoon, Shin, Hwang, and Yang]{yoon2021fedmix}
Tehrim Yoon, Sumin Shin, Sung~Ju Hwang, and Eunho Yang.
\newblock Fedmix: Approximation of mixup under mean augmented federated learning.
\newblock In \emph{International Conference on Learning Representations}, 2021.
\newblock URL \url{https://openreview.net/forum?id=Ogga20D2HO-}.

\bibitem[Yu et~al.(2020)Yu, Bagdasaryan, and Shmatikov]{yu2020salvaging}
Tao Yu, Eugene Bagdasaryan, and Vitaly Shmatikov.
\newblock Salvaging federated learning by local adaptation.
\newblock \emph{arXiv preprint arXiv:2002.04758}, 2020.

\bibitem[Yuan et~al.(2022)Yuan, Morningstar, Ning, and Singhal]{yuan2022what}
Honglin Yuan, Warren~Richard Morningstar, Lin Ning, and Karan Singhal.
\newblock What do we mean by generalization in federated learning?
\newblock In \emph{International Conference on Learning Representations}, 2022.
\newblock URL \url{https://openreview.net/forum?id=VimqQq-i_Q}.

\bibitem[Zhang et~al.(2023)Zhang, Hua, Wang, Song, Xue, Ma, and Guan]{zhang2023fedala}
Jianqing Zhang, Yang Hua, Hao Wang, Tao Song, Zhengui Xue, Ruhui Ma, and Haibing Guan.
\newblock Fedala: Adaptive local aggregation for personalized federated learning.
\newblock In \emph{Proceedings of the AAAI Conference on Artificial Intelligence}, volume~37, pages 11237--11244, 2023.

\bibitem[Zhang et~al.(2021)Zhang, Guo, Ma, Wang, Xu, and Wu]{zhang2021parameterized}
Jie Zhang, Song Guo, Xiaosong Ma, Haozhao Wang, Wenchao Xu, and Feijie Wu.
\newblock Parameterized knowledge transfer for personalized federated learning.
\newblock \emph{Advances in Neural Information Processing Systems}, 34:\penalty0 10092--10104, 2021.

\bibitem[Zhao et~al.(2018)Zhao, Li, Lai, Suda, Civin, and Chandra]{zhao2018federated}
Yue Zhao, Meng Li, Liangzhen Lai, Naveen Suda, Damon Civin, and Vikas Chandra.
\newblock Federated learning with non-iid data.
\newblock \emph{arXiv preprint arXiv:1806.00582}, 2018.

\end{thebibliography}
\bibliographystyle{plainnat}
}

\clearpage

\appendix

\section{Implementation Details.} 
\label{app:impl_details}
In our methods, we adopt the SGD optimizer with a learning rate of 0.1. We set the number of local epochs $E = 5$, and batch size equals 100, consistent with the setup in \citep{wu2023bold}. The maximum communication rounds $T$ is set to 300 to ensure full convergence. 
The same settings are applied to other baseline methods for fair comparison. In each run, we evaluate the uniform averaging test accuracy across all clients in each communication round and select the final accuracy as the final result. Each experiment is repeated three times under different seeds, and the mean and standard deviation are reported.

\section{Sensitivity-based Criterion Derivation}
\label{appendix:sensitivity}

In this section, we derive the sensitivity criterion used in LIPS for evaluating the importance of each parameter during federated training.

\textbf{Definition of sensitivity.}
Given a model \( \mathbf{w}_i^t \) at client \( i \) during communication round \( t \), the parameter set is expressed as:
\[
\Theta = \{w_{i,1}^t, w_{i,2}^t, \ldots, w_{i,j}^t, \ldots, w_{i,n}^t\}.
\]
The sensitivity of the \( j \)-th parameter \( w_{i,j}^t \) is defined as the absolute difference in the loss function \( \mathcal{L} \) when the parameter \( w_{i,j}^t \) is zeroed out:
\begin{equation}
s_{i,j}^t = \left| \mathcal{L}(\Theta) - \mathcal{L}(w_{i,1}^t, \ldots, w_{i,j-1}^t, 0, w_{i,j+1}^t, \ldots, w_{i,n}^t) \right|,
\end{equation}
where \( \mathcal{L} \) is the task-specific loss function. This sensitivity metric quantify the importance of the parameter \( w_{i,j}^t \) to the local task, with higher sensitivity values indicating parameters that are more critical to the model’s performance.

\textbf{Taylor approximation.}
Directly computing \( s_{i,j}^t \) for each parameter requires additional forward passes, significantly increasing computational costs. To overcome this, we use a first-order Taylor approximation to approximate the sensitivity:
\begin{equation}
s_{i,j}^t = \left| \mathcal{L}(\Theta) - \mathcal{L}(w_{i,1}^t, \ldots, 0, \ldots, w_{i,n}^t) \right| \approx \left| \nabla_{w_{i,j}} \mathcal{L}(\Theta) \cdot w_{i,j}^t + R_1(\Theta) \right|,
\end{equation}
where \( R_1(\Theta) \) represents higher-order terms that are neglected due to the linear approximation. Thus, the sensitivity can be approximated as:
\begin{equation}
s_{i,j}^t \approx \left| \nabla_{w_{i,j}} \mathcal{L}(\Theta) \cdot w_{i,j}^t \right|.
\end{equation}
This approximation requires only a single backpropagation pass to compute the gradients, significantly reducing computational overhead.

\textbf{Incorporating local updates in federated learning.}
In federated learning, each client performs multiple local training epochs before sending updates to the server. To incorporate this, we replace \( \nabla_{w_{i,j}} \mathcal{L} \) with the variation in the parameter \( w_{i,j}^t \) over local training epochs. Specifically:
\begin{equation}
\Delta w_{i,j}^t = w_{i,j}^{t} - w_{i,j}^{t'},
\end{equation}
where \( w_{i,j}^{t'} \) and \( w_{i,j}^{t} \) denote the parameter values before and after the local training update, respectively. Substituting this into the sensitivity definition, we obtain:
\begin{equation}
s_{i,j}^t = \left| \Delta w_{i,j}^t \cdot w_{i,j}^{t} \right|.
\end{equation}

\textbf{Sensitivity-based metric}
The final sensitivity metric for parameter \( w_{i,j}^t \) at client \( i \) during round \( t \) is given by:
\begin{equation}
s_{i,j}^t = \left| \Delta w_{i,j}^t \cdot w_{i,j}^{t} \right|,
\end{equation}
where \( \Delta w_{i,j}^t = w_{i,j}^{t} - w_{i,j}^{t'} \).

This metric captures both the magnitude of local updates (\( \Delta w_{i,j}^t \)) and the parameter's final state (\( w_{i,j}^{t} \)), making it well-suited for identifying less critical parameters that can be zeroed out to enforce sparsity while preserving model performance.

\section{Additional Layer-wise Cosine Similarity Visualization}
\label{app:cosine}

\begin{figure}[!htb]
    \centering
    \includegraphics[width=\textwidth]{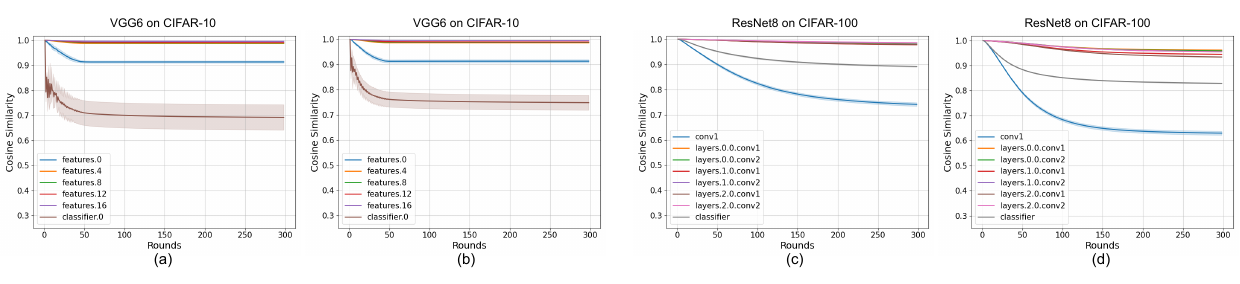}
    \caption{Layer-wise cosine similarity during the training process. We track the global model's layer-wise cosine similarity relative to the 2nd communication round states after aggregation throughout training on CIFAR-10 and CIFAR-100 datasets using VGG6 and ResNet-8 architectures, respectively. The experiments are conducted with 100 clients, each having 100 training samples. For CIFAR-10, results are shown for (a) Dir($\alpha$=0.5) and (b) Dir($\alpha$=1.0), while for CIFAR-100, results are shown for (c) Dir($\alpha$=0.01) and (d) Dir($\alpha$=0.5). The legend indicates the layer names for each architecture.}
    \label{fig: app_inertia}
\end{figure}

In Figure \ref{fig: app_inertia}, we present additional results on the layer-wise cosine similarity of the global model after aggregation throughout training under different federated learning settings for CIFAR-10 and CIFAR-100. Consistently, we observe that most layers, particularly the middle layers, exhibit minimal changes during the training process after the early stages of training. This is evidenced by the consistently high cosine similarity values (above 0.95) for the middle layers' weights when compared to their states in the early training phase. These results confirm that this behavior is consistently observed across different datasets and architectures.

\begin{figure}[!htb]
    \centering
    \includegraphics[width=\textwidth]{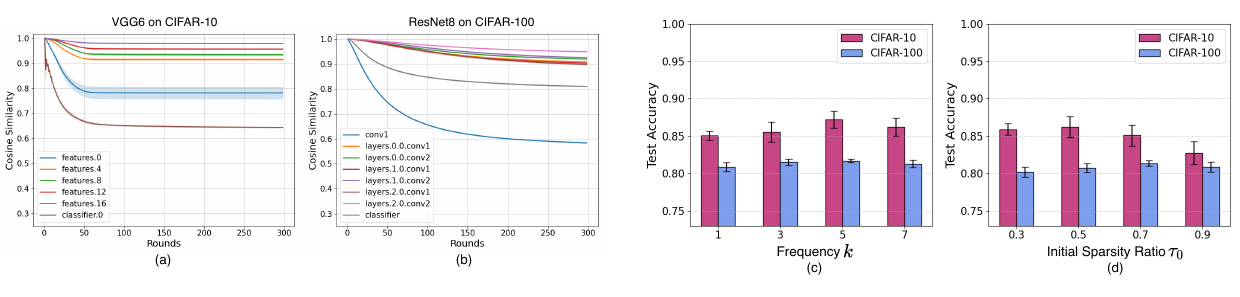}
    \vskip -0.15in
    \caption{Layer-wise cosine similarity during the training process. (a) and (b) show the global model's layer-wise cosine similarity relative to the 2nd communication round for CIFAR-10 and CIFAR-100 datasets with 300 training samples per client, under Dir($\alpha$=0.5) and Dir($\alpha$=0.01), respectively. (c) illustrates the effect of different sparsification intervals $k$, showing how the frequency of introducing sparsity impacts performance. (d) highlights the impact of varying initial sparsity ratios $\tau_0$ on the performance of CIFAR-10 (Dir($\alpha$=0.1) and CIFAR-10 (Dir($\alpha$=0.01)).}
    \label{fig: app_ablation}
\end{figure}

Additionally, we provide layer-wise cosine similarity results for the global model throughout training on CIFAR-10 and CIFAR-100 under a scenario with 300 training samples per client. As highlighted in our main claim, fewer training samples exacerbate the layer-wise inertia phenomenon. From Figure \ref{fig: app_ablation} (a) and (b), we observe that with more training samples per client, the layer-wise inertia phenomenon is notably weakened compared to the scenario with 100 training samples per client, as shown in Figure \ref{fig: app_inertia} (a) and (c), respectively. These observations further support our claim in Section \ref{inertia_phenomenon}.

\section{Layer-wise Learning Behavior in Low-Data Centralized Training}
\label{app:centralize}

\begin{figure}[!htb]
    \centering
    \vskip -0.1in
    \includegraphics[width=\textwidth]{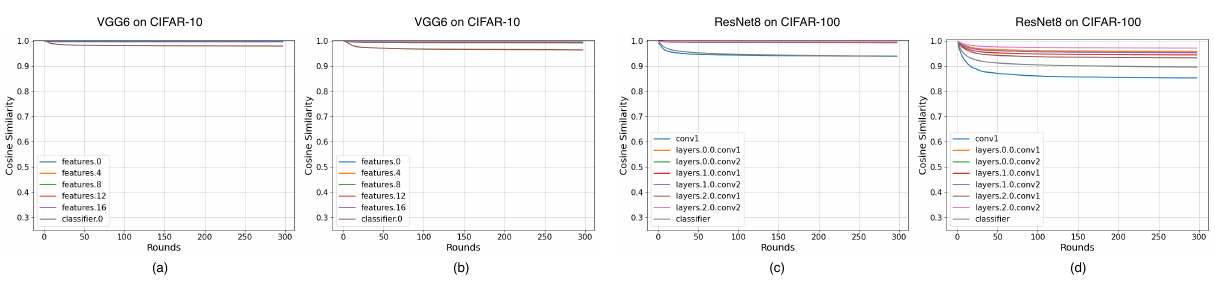}
    \vskip -0.15in
    \caption{Layer-wise cosine similarity of models under centralized training throughout the training process. We track the cosine similarity of each layer in the global model relative to its state at the 2nd communication round, using CIFAR-10 and CIFAR-100 datasets with VGG6 and ResNet-8 architectures, respectively. Experiments are conducted with 100 clients under Dir($\alpha$=0.1), where each client holds 100 training samples for subfigures (a) and (c), and 300 samples for (b) and (d). The legend indicates the corresponding layer names for each architecture.}
    \label{fig: central_inertia}
    \vskip -0.1in
\end{figure}

To further understand the origins of the Layer-wise Inertia Phenomenon, we conduct layer-wise learning behavior in low-data centralized training settings, as shown in Figure \ref{fig: central_inertia}. We can observe that inertia is also evident in centralized training when data is scarce, suggesting that this phenomenon primarily arises from overfitting due to insufficient training data, which results in minimal parameter updates in certain layers.

Interestingly, we observe that FL offers partial mitigation. Specifically, the first and last layers, typically more sensitive to input variation and task-specific supervision, show lower cosine similarity across communication rounds in FL compared to centralized training, indicating more effective adaptation, compared to Figure \ref{fig: fd_inertia}. However, for intermediate layers, the cosine similarity remains consistently high in both settings. While FL provides slight improvements, it still struggles to significantly alter the stagnant behavior of these layers, highlighting a shared challenge across both training paradigms. These findings underscore the need for targeted solutions, such as our proposed sparsity-based strategy, to revitalize underutilized layers in low-data regimes.

\section{Additional Visualization of the Impact of LIPS on Layer Dynamics}
\label{app:visual}

To further validate the effectiveness of LIPS in addressing the Layer-wise Inertia Phenomenon, we provide additional visualizations across different datasets and model architectures beyond those presented in the main text.

\begin{figure}[!htb]
    \centering
    \includegraphics[width=\textwidth]{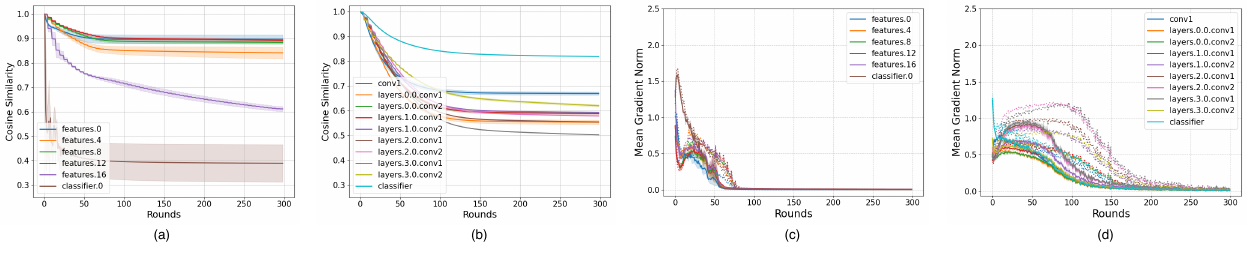}
    \vskip -0.15in
    \caption{Layer-wise cosine similarity and gradient norm analysis during the training process. (a) and (b) show the global model's layer-wise cosine similarity relative to the 2nd communication round for CIFAR-10 and TinyImageNet datasets, with Dir($\alpha$=0.1) and Dir($\alpha$=0.01), respectively. (c) and (d) show comparison of layer-wise mean gradient norms across all clients between FedBN (solid line) and LIPS (dotted line) after each training round for CIFAR-10 and TinyImageNet datasets, with Dir($\alpha$=0.1) and Dir($\alpha$=0.01), respectively.}
    \label{fig: app_visual_grad}
\end{figure}

In Figure~\ref{fig: app_visual_grad}, we extend our analysis of layer-wise cosine similarity and mean gradient norm to CIFAR-10 and TinyImageNet datasets using the VGG6 and ResNet-10 architectures, respectively. Specifically, subfigures (a) and (b) show the global model's layer-wise cosine similarity relative to its 2nd round checkpoint under Dirichlet data distributions with $\alpha=0.1$ and $\alpha=0.01$ for CIFAR-10 and TinyImageNet datasets, respectively. These results confirm that LIPS reduces cosine similarity in intermediate layers over the course of training, indicating more dynamic updates and better mitigation of stagnation.

Subfigures (c) and (d) further compare the mean gradient norm per layer between FedBN and LIPS across training rounds for CIFAR-10 and TinyImageNet datasets. LIPS consistently increases the gradient magnitudes, particularly in the middle layers, thereby promoting more effective local updates. This sustained update activity contributes to better and more meaningful collaboration through aggregation in FL environments.

These visualizations reinforce the generality of LIPS across diverse federated learning scenarios and demonstrate its robustness in enhancing learning dynamics across multiple datasets and model backbones.

\section{Ablation Study}
\label{app:ablation}

In this section, we perform a comprehensive ablation study to investigate the impact of two key hyperparameters in our method: the sparsification frequency 
$k$ and the initial sparsity ratio $\tau_0$. Below, we detail the effect of these parameters and highlight the configurations that yield optimal performance in our approach.

\subsection{Effect of frequency $k$}
In Figure \ref{fig: app_ablation} (c), we evaluate the performance of our method under varying sparsification intervals $k$, which determine how frequently sparsity is introduced during communication rounds. The highest accuracies are achieved when sparsification is performed every 5 rounds in non-IID settings with Dir($\alpha$=0.1) for CIFAR-10 and Dir($\alpha$=0.01) for CIFAR-100, a setting adopted as the default in our main experiments. 
The findings suggest that overly frequent sparsification can disrupt training convergence by introducing excessive instability in the model's sparse topology. On the other hand, infrequent sparsification reduces its effectiveness in addressing the Layer-wise Inertia phenomenon, as the model fails to sufficiently adapt its sparse topology during training. Striking the right balance in sparsification frequency is therefore essential for optimizing performance in our settings.

\subsection{Effect of initial sparsity ratio $\tau_0$} 
To assess the impact of LIPS under different initial sparsity ratios, which determine the proportion of weights pruned from specific layers in each round, we conducted experiments illustrated in Figure \ref{fig: app_ablation} (d). The results reveal that the optimal sparsity ratio varies across datasets, as tasks with differing complexities and models of varying scales exhibit distinct levels of weight redundancy.
In our experiments, an initial sparsity ratio of 0.5 was most effective for CIFAR-10 with the VGG6 architecture. For CIFAR-100 with ResNet models, a higher initial sparsity ratio of 0.7 yielded better results. This can be attributed to the deeper architecture and residual connections of ResNet, which might provide greater representational power and allow the model to maintain performance even under higher sparsity levels.

\section{Performance Comparison with Smaller Models}

As analyzed in Section \ref{inertia_phenomenon}, having more layers exacerbates the Layer-wise Inertia phenomenon. Conversely, fewer layers help mitigate this issue. This raises an intuitive question: \textit{why not simply use smaller models in such cases?}

\begin{wraptable}{r}{0.55\textwidth}
\begin{minipage}{0.55\textwidth}
\vskip -0.1in
\caption{Performance comparison of FedBN and LIPS on CIFAR-100 using various ResNet-based architectures: ResNet6, ResNet8, and ResNet10, under data distribution Dir($\alpha$=0.1) with 100 clients, each having 100 training samples.}
\resizebox{\linewidth}{!}{
\begin{tabular}{c|c|c|c}
\toprule[1.5pt]
{Method} & {ResNet6} & ResNet8 & ResNet10 \\
\hline
\bottomrule[1pt]
{FedBN} & {40.44$\pm$0.41} & {43.08$\pm$0.45} & {42.84$\pm$0.19} \\
{LIPS} & {44.56$\pm$0.38} & {47.84$\pm$0.47} & {47.54$\pm$0.32} \\
\bottomrule[1.5pt]
\end{tabular} 
}
\label{tab: app_layers}
\end{minipage}
\end{wraptable}%

To address this concern, we compare the final performance of smaller models with models having more layers in Table \ref{tab: app_layers}. Specifically, we evaluate ResNet6, ResNet8, and ResNet10 on CIFAR-100 with 100 training samples per client under a Dirichlet data distribution Dir($\alpha$=0.1). The results show that while ResNet6 reduces the Layer-wise Inertia phenomenon, it performs worse overall compared to ResNet8 and ResNet10. The primary reason is that smaller models lack the learning capacity required to handle the complexity of the task, resulting in inferior performance.
Therefore, addressing the Layer-wise Inertia phenomenon is crucial to unlocking the full potential of larger models, enabling them to perform optimally in federated learning scenarios.

\section{Performance Comparison with Local Sparse Training}
\label{app:sparse_training}

In this section, we compare the performance of LIPS with and without maintaining sparsity throughout local training, as shown in Table~\ref{tab:local_sparse}. The $\tt w/.\hspace{0.2em}local \hspace{0.2em} sparsity$ setting enforces sparsity during both the global aggregation phase and local updates, while the $\tt w/o.\hspace{0.2em}local \hspace{0.2em} sparsity$ configuration uses transient sparsity—applying sparsity only after aggregation and lifting it before local training.

The results show that maintaining sparsity throughout training ($\tt w/.\hspace{0.2em}local \hspace{0.2em} sparsity$) leads to slightly lower accuracy compared to the transient sparsity variant ($\tt w/o.\hspace{0.2em}local \hspace{0.2em} sparsity$); however, the performance difference is modest and both settings still outperform FedBN, which does not incorporate sparsity at all. For example, on CIFAR-100 with $\alpha=0.5$, accuracy drops from 28.31\% to 26.64\%. However, this comes with a substantial benefit: applying sparsity only at aggregation (transient sparsity) reduces training computation by nearly 2×, since dense updates are used locally. This highlights a practical trade-off between training efficiency and performance. While full sparse training can further reduce computation cost, the transient sparsity mechanism used in LIPS retains dense local training, which preserves model quality more effectively.

\begin{table*}[!htbp]
\centering
\caption{Performance comparison on CIFAR-100, and TinyImageNet using different architectures (ResNet-8, and ResNet-10, respectively), with 100 clients under varying values of $\alpha$. We report both accuracy(\%) ($\uparrow$) and training FLOPs($10^{12}$) ($\downarrow$) under initial sparsity ratio $\tau_0=0.7$.}
\label{tab:local_sparse}
\resizebox{\linewidth}{!}{
\begin{tabular}{l|ccc|ccc}
\toprule[1.5pt]
\multirow{2}{*}{Method} & \multicolumn{3}{c|}{CIFAR-100} & \multicolumn{3}{c}{TinyImageNet} \\ \cline{2-7} 
 & $\alpha=0.1$ & $\alpha=0.5$ & FLOPs & $\alpha=0.1$ & $\alpha=0.5$ & FLOPs \\ \hline
\bottomrule[1pt]
FedBN & 43.08$\pm$0.45 & 24.45$\pm$0.61 &14.9 & 36.83$\pm$0.22 & 20.73$\pm$1.23 & 219.4\\ 
w/o. local sparsity & 47.96$\pm$0.54 & 28.31$\pm$0.54 &1.0× & 40.50$\pm$0.32 & 23.50$\pm$0.53 &1.0× \\
w/. local sparsity & 46.26$\pm$0.46 & 26.64$\pm$0.62 &0.6× & 38.92$\pm$0.45 & 22.18$\pm$0.43 &0.6× \\

\bottomrule[1.5pt]
\end{tabular}
}
\vskip -0.1in
\end{table*}

\section{Discussion}
\label{app:discussion}

\subsection{Dropout and Other Regularization Techniques}

While our method shares the high-level objective of mitigating overfitting with techniques like dropout, it differs significantly in motivation, mechanism, and effectiveness in federated learning (FL). Dropout typically deactivates neurons randomly during training to prevent co-adaptation, resetting at every iteration. In contrast, our approach introduces structured sparsity by selectively pruning weights based on sensitivity, maintaining these sparsity patterns throughout training.

Importantly, our method is grounded in the observed Layer-wise Inertia phenomenon, targeting layers that suffer from stagnation due to overfitting—something dropout does not explicitly address. Additionally, our pruning mechanism is more flexible: it can be combined with various initialization strategies (such as original weight reinitialization), offering further performance benefits, as shown in our ablation studies \ref{sec:ablation}.

While dropout has been applied in FL (e.g., for personalization or regularization purposes \citep{wen2022federated, jeon2023federated}), it does not explicitly aim to tackle the challenge of persistent parameter inactivity in intermediate layers of global models. 
In contrast, our approach is driven by a novel empirical observation—the Layer-wise Inertia Phenomenon—which reveals the stagnation of certain layers during training in low-data regimes. By directly targeting this issue, our method moves beyond generic regularization by introducing sensitivity-guided sparsity that selectively reactivates and stimulates underutilized layers. This design makes our approach not only more principled and interpretable, but also particularly effective in enhancing aggregation and collaboration in federated learning, especially when training data is limited.

\subsection{Existing Sparsity Techniques in FL}

Sparsity has primarily been used in FL to reduce communication overhead, such as through gradient sparsity \citep{lu2024data, wangni2018gradient} or model pruning \citep{jiang2022model, jiang2023complement}. These techniques target communication or storage efficiency, whereas LIPS is designed to address a fundamentally different challenge: improving learning dynamics and collaboration in low-data regimes. Recent works have considered dynamic sparse training in centralized settings \citep{mocanu2018scalable, evci2020rigging}, and some efforts extend such ideas to FL \citep{bibikar2022federated, jiang2022model, chen2023efficient}. However, they often focus on training efficiency or latency rather than tackling representation stagnation due to limited updates in global aggregation. Our method complements these efforts by offering a simple yet effective way to reinvigorate stagnant layers without additional communication overhead.

\section{Limitations and Future Work} 
\label{app:limits}


While LIPS targets the middle layers to address inertia, extending the sparsification to other layers, including the first and last layers, could be beneficial under certain conditions. This would require exploring layer-specific sensitivities to sparsity and determining their impact on both model performance and stability during global aggregation.

The simplicity of LIPS also makes it a strong candidate for integration into existing FL frameworks with minimal changes. Future enhancements could involve combining LIPS with more sophisticated aggregation techniques, such as weighted averaging based on client contributions, or integrating personalization strategies that allow clients to maintain task-specific models while benefiting from global knowledge. These combinations have the potential to further improve collaboration and enhance performance in heterogeneous federated environments.

Lastly, the current evaluation of LIPS has been limited to specific datasets and architectures. Expanding the scope to include more diverse datasets in other domains, larger-scale models, and real-world federated learning applications would provide a broader understanding of its effectiveness and generalizability. These future directions highlight the potential for LIPS to evolve into a more comprehensive and scalable solution for efficient and effective federated learning.

\section{Impact Statements} 
\label{app:impact}

This paper contributes to the understanding and optimization of federated learning by analyzing layer-wise learning dynamics. Our findings provide valuable insights into federated optimization, potentially improving real-world applications where communication efficiency and model adaptability are essential. Although our contributions do not inherently lead to negative societal impacts, we encourage the community to remain mindful of potential implications when extending our research.

\end{document}